\documentclass{article}

\PassOptionsToPackage{numbers, compress}{natbib}

 \usepackage[preprint]{neurips_2026}

\usepackage[utf8]{inputenc} % allow utf-8 input
\usepackage[T1]{fontenc}    % use 8-bit T1 fonts
\usepackage{hyperref}       % hyperlinks
\usepackage{url}            % simple URL typesetting
\usepackage{booktabs}       % professional-quality tables
\usepackage{amsfonts}       % blackboard math symbols
\usepackage{nicefrac}       % compact symbols for 1/2, etc.
\usepackage{microtype}      % microtypography
\usepackage[table,xcdraw]{xcolor}         % colors

\usepackage{amsmath}
\usepackage{graphicx}
\usepackage{subcaption}
\usepackage{placeins}
\usepackage{makecell}
\usepackage{multirow}
\usepackage{tabularx}
\usepackage{cleveref}
\usepackage{wrapfig}
\usepackage{adjustbox}

\title{Video as Natural Augmentation: Towards\\ Unified AI-Generated Image and Video Detection}

\author{%
  Zhengcen Li\textsuperscript{1}, Chenyang Jiang\textsuperscript{1}, Liangxu Su\textsuperscript{1}, Tong Shao\textsuperscript{1}, Shiyang Zhou\textsuperscript{1, 3} \\
  \textbf{Ming Tao\textsuperscript{2}, Jingyong Su\textsuperscript{1}} \\
  \textsuperscript{1} Harbin Institute of Technology, Shenzhen\\
  \textsuperscript{2} Pengcheng Laboratory\\
  \textsuperscript{3} Shenzhen Loop Area Institute\\
  \texttt{lizhengcen@stu.hit.edu.cn, sujingyong@hit.edu.cn}
}

\begin{document}

\maketitle

\begin{abstract}
AI-generated content (AIGC) is rapidly improving, creating an urgent need for detectors that generalize across data sources, deployment pipelines, and visual modalities. A strongly generalizable detector should remain robust under distributional variations. However, we identify a consistent failure mode: SOTA AI-generated image detectors often collapse when applied to frames extracted from videos. Through systematic analysis, we show that this cross-modal gap arises from both entangled \emph{synthesis-agnostic} video processing shifts, including color conversion, codec compression, resizing, and blur, and \emph{model-specific} fingerprints introduced by modern video generators. Motivated by these findings, we propose \textbf{VINA} (\textbf{VI}deo as \textbf{N}atural \textbf{A}ugmentation), a unified AIGC detection framework that jointly trains on image and video data. VINA uses video frames as physically grounded natural augmentations and further introduces a cross-modal supervised contrastive objective to align image and video representations under a shared real/fake decision boundary. Extensive experiments on 14 image, video, and in-the-wild benchmarks show that VINA delivers bidirectional gains, improves robustness and transferability, and achieves state-of-the-art performance across nearly all evaluated settings without complex augmentation or dataset-specific tuning.
\end{abstract}

\section{Introduction}

Recent advances in AI-generated content (AIGC) have enabled diffusion models and large-scale transformers to produce highly realistic and semantically consistent images~\cite{arxiv2024stablediffusion3} and videos~\cite{sora}. As generation quality improves, synthetic media increasingly blurs the boundary between authentic and generated content, lowering the cost of creating persuasive misinformation at scale and weakening public trust~\cite{wang2024aigcsecurity}. The broad availability of both image and video generation models and products therefore creates an urgent demand for accurate, robust, and broadly generalizable detectors that can operate reliably in diverse real-world settings.

\begin{figure}[t]
    \centering
    \begin{minipage}[c]{0.43\linewidth}
        \centering
        \includegraphics[width=\linewidth]{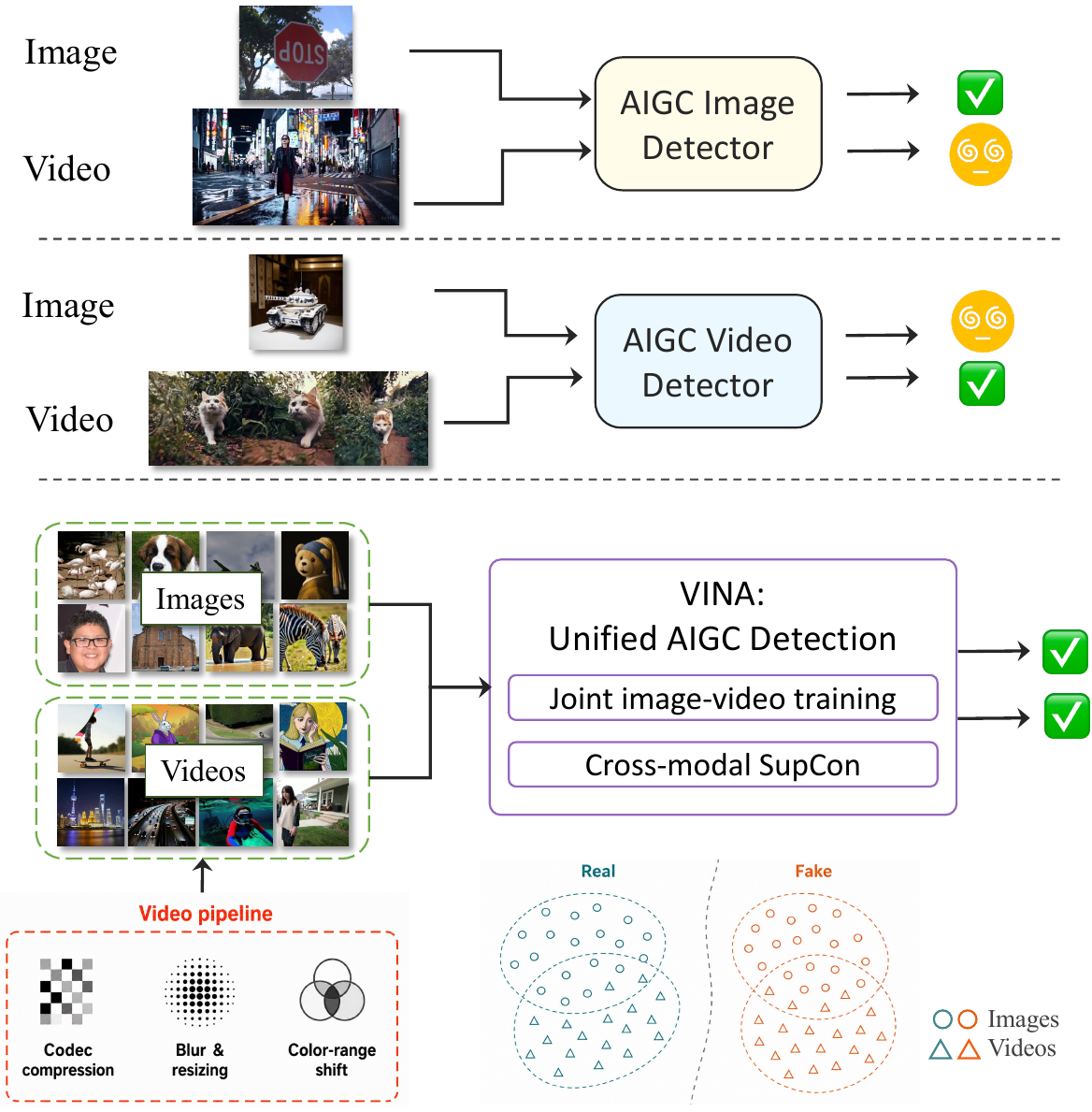}
        \phantomsubcaption\label{fig:intro:left}
    \end{minipage}
    \hspace{0.035\linewidth}
    \begin{minipage}[c]{0.43\linewidth}
        \centering
        \includegraphics[width=\linewidth]{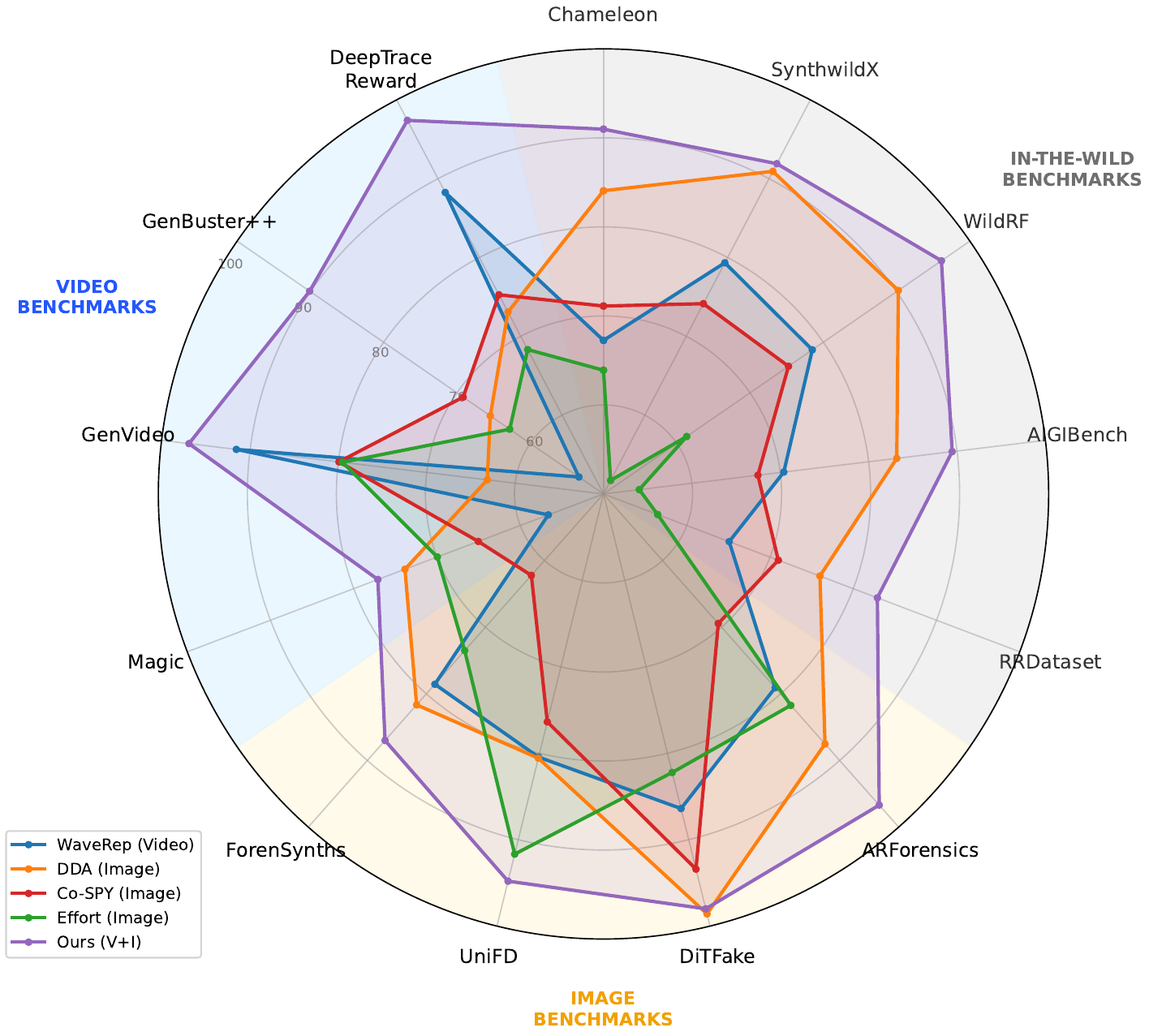}
        \phantomsubcaption\label{fig:intro:right}
    \end{minipage}
    \caption{\textbf{Motivation and benchmark performance of unified AIGC detection.} \textbf{Left:} Asymmetric cross-modal failures motivate VINA to learn from both image and video data. \textbf{Right:} Average accuracy across image, video, and in-the-wild AIGC benchmarks shows that image-based detectors degrade sharply on videos, while joint training improves cross-modal generalization.}
    \label{fig:intro}
    \vspace{-5mm}
\end{figure}

Generalization is the central challenge in AIGC detection. A practical detector must transfer across generators, data sources, and deployment pipelines, especially to in-the-wild media whose provenance and post-processing history are unknown. Existing AI-generated image (AIGI) detectors mainly exploit local pixel artifacts~\cite{cvpr2024npr}, frequency-domain traces~\cite{kdd2025safe}, reconstruction behavior~\cite{cvpr2025co-spy}, or high-level semantics~\cite{cvpr2023unifd, icml2025effort}. Recent work further improves generalization through data alignment~\cite{nips2025dda}, where diffusion or VAE models inject controlled synthetic traces into real images at the pixel~\cite{cvpr2025b-free} or frequency level~\cite{nips2025seeingwhatmatters}. Although these methods improve cross-generator and cross-dataset transfer, their performance still deteriorates under unseen generators~\cite{eccv2024zeroshot}, real-world online samples~\cite{iclr2025sanity}, and media subjected to heavy compression or platform degradation~\cite{iccv2025rrdataset}. This indicates that current detectors remain vulnerable to shortcut cues and distribution shifts that are not tied to synthesis itself.

In this paper, we identify a novel systematic failure mode: state-of-the-art AIGI detectors collapse when evaluated on video frames. As summarized in the left and right panels of \Cref{fig:intro}, image-trained detectors that perform competitively on image benchmarks often drop sharply on AI-generated video (AIGV) benchmarks, with several methods approaching random guessing. This failure persists even for recent detectors designed for stronger cross-generator transfer, showing that the problem is not merely a matter of unseen generators. Meanwhile, dedicated AIGV detectors~\cite{arxiv2024demamba,nips2025nsg-vd,wang2023allseeing} remain relatively underexplored and often fail to preserve strong image-domain generalization. In practice, a AIGC detector cannot assume its input is a clean image; it may receive a frame extracted from a compressed video or an image after online platform degradation. These observations suggest that treating AIGI and AIGV detection as separate tasks leaves a fundamental cross-modal generalization gap.

We analyze this failure from the properties of real video data. Video frames are not simply still images sampled from another generator distribution. They are produced by real acquisition, rendering, compression, and distribution pipelines. Compared with common image datasets, video frames contain stronger \emph{synthesis-agnostic} shifts, including realistic imaging and motion blur, codec-induced frequency decay, limited-range color conversion, resizing, and repeated platform recompression. They also cover more diverse scenes and camera motions, while modern video generators introduce additional model-specific fingerprints related to temporal modules and motion synthesis. These factors suppress high-frequency micro-artifacts that image detectors often exploit as shortcuts, while introducing low- and mid-frequency statistics that are underrepresented in image-only training. We therefore view video frames as physically grounded natural augmentations: they expose detectors to realistic degradations that are difficult to reproduce with handcrafted JPEG, blur, or color perturbations.

To address this gap, we propose \textbf{VINA} (\textbf{VI}deo as \textbf{N}atural \textbf{A}ugmentation), a joint video--image training framework for unified AIGC detection. VINA augments existing image training sets~\cite{cvpr2020cnngenerated,nips2023genimage,nips2025dda} with AIGV detection data~\cite{arxiv2024demamba,iclr2026preserving} and trains a single detector for both image and video inputs. Beyond binary supervision on mixed batches, we introduce a cross-modal supervised contrastive (CM-SupCon) loss that aligns image and video features with the same real/fake label while preserving the decision boundary. This regularizes the unified representation space, reduces reliance on modality- or source-specific shortcuts, and encourages the detector to use generative traces that remain discriminative across both clean images and degraded video frames. Extensive experiments show bidirectional gains: joint training substantially improves video detection while also improving image and in-the-wild performance. Without complex augmentation, explicit alignment, or dataset-specific tuning, VINA achieves strong state-of-the-art results across multiple backbones and data-source combinations, reaching 90.4\% average accuracy on video benchmarks, 94.1\% on image benchmarks, and 92.9\% on in-the-wild benchmarks. Our contributions are summarized as follows.

\begin{enumerate}
    \item We reveal that state-of-the-art AIGI detectors systematically fail on video benchmarks, formulate unified AIGC detection as a generalization problem, and show that training and evaluation should jointly cover AI-generated images and videos rather than treating them as isolated tasks.
    \item We propose VINA, a joint video--image training framework with a CM-SupCon objective that regularizes image and video representations under a unified decision boundary, suppressing source- and modality-specific shortcut cues.
    \item Extensive experiments across 14 image, video, and in-the-wild benchmarks show that VINA brings bidirectional gains for both videos and images, substantially improves in-the-wild robustness, and generalizes across multiple backbones and data sources without complex augmentation, explicit alignment, or dataset-specific tuning.
\end{enumerate}

\section{Related Work}
\label{sec:related_work}

\subsection{Generative Models} 

\paragraph{Image Generative Models.} Image generation has evolved from GAN-based models~\cite{nips2014gan} to VAE/VQ-VAE frameworks~\cite{arxiv2022autoencoding,arxiv2018vq-vae} and diffusion models~\cite{nips2020ddpm}. Latent Diffusion Models (LDMs)~\cite{cvpr2022stablediffusion} improve efficiency by denoising in a compressed latent space, while recent Diffusion Transformers (DiTs)~\cite{iccv2023dit}, autoregressive models~\cite{nips2024var} and unified models~\cite{arxiv2025bagel} further improve scalability with transformer-based architectures.

\paragraph{Video Generative Models.} Video generation extends image generation to the spatiotemporal domain. Early systems such as Make-A-Video~\cite{arxiv2022make-a-video} and Stable Video Diffusion~\cite{arxiv2023stablevideodiffusion} adapt image diffusion backbones with temporal modeling. Recent representative systems, including Sora~\cite{sora}, Wan~\cite{arxiv2025wan}, Kling~\cite{kling}, and Seedance~\cite{arxiv2025seedance1.0}, increasingly rely on video diffusion transformers, multimodal conditioning, and large-scale pretraining and post-training to improve visual fidelity, motion consistency, and controllability.

Different generative architectures often leave distinct generative traces. To ensure systematic evaluation across modalities, we assemble a comprehensive detection benchmark that spans the major image and video architectures, including GAN, LDM, DiT, autoregressive models, as well as challenging in-the-wild samples.

\subsection{AIGC Detection}
\paragraph{AI-Generated Image Detection.} 
Significant research has focused on developing generalizable methods for detecting synthetic images~\cite{cvpr2020cnngenerated,cvpr2023unifd}. These include approaches based on spectral artifacts~\cite{kdd2025safe}, reconstruction error~\cite{iccv2023dire,icml2024drct,iccv2025d^3qe}, pixel-level features~\cite{cvpr2024npr,cvpr2025co-spy}, and the adaptation of visual backbones~\cite{cvpr2024fatformer,eccv2024rine,icml2025effort,aaai2025c2p-clip}.  Typically trained on images from specific generative models~\cite{iclr2018progan,nips2023genimage}, these methods aim to achieve generalization across both different models and architectures.

\paragraph{AI-Generated Video Detection.} Early methods for detecting AI-generated videos (AIGVs) primarily targeted facial manipulations or biological inconsistencies. With the advent of high-fidelity video synthesis, the focus has shifted toward detecting fully AI-generated scenes~\cite{cvpr2025unite}. This progress is supported by million-scale benchmarks such as GenVideo~\cite{arxiv2024demamba} and GenVidBench~\cite{arxiv2024genvidbench}, which evaluate cross-generator generalization. Physics-driven methods like D3~\cite{iccv2025d3} and NSG-VD~\cite{nips2025nsg-vd} identify synthetic content by quantifying deviations from Newtonian mechanics or violations of probability flow conservation in the spatiotemporal latent space. Recent video-based methods~\cite{iclr2026preserving,arxiv2025vidguard-r1,arxiv2025skyra,arxiv2026videoveritas} or unified methods~\cite{arxiv2025busterx++,arxiv2025loki,arxiv2025ivy-fake} use multimodal large language models to provide human-interpretable reasoning and artifact localization, but they remain limited by LLM hallucination and high computational cost; moreover, their vision encoders are primarily optimized for high-level semantic perception and may be insufficiently sensitive to subtile artifacts~\cite{arxiv2026aligngemini}.

\paragraph{Dataset Alignment.} Relying exclusively on frequency-domain cues or on training data tied to a specific generative model can introduce vulnerability to dataset bias~\cite{nips2025dda}, thereby limiting generalization. To address this, recent works employ generative models such as diffusion models~\cite{icml2024drct} or VAEs~\cite{iclr2025aligned,arxiv2025beyondartifacts} to inject controlled artifact patterns, either in pixel space~\cite{cvpr2025b-free} or in the frequency domain~\cite{nips2025seeingwhatmatters}. By strictly isolating these injected artifacts from other confounding factors, such datasets facilitate the training of more robust and generalizable detection models.

However, existing methods are typically confined to a single task, focusing either on image or video detection, and do not account for the intrinsic differences between still images and video frames. In contrast, we explicitly unify these two tasks within a single training framework to enhance the generalization capability of visual AIGC detection.

\section{Method}

\subsection{Motivation and Analysis}
\label{sec:method_analysis}

\paragraph{Why do AIGI detectors fail on video frames?}

Despite semantic similarities between generated images and video frames, our empirical observations in the right panel of \cref{fig:intro} reveal a significant performance drop when state-of-the-art AI-generated image (AIGI) detectors are applied to video frames. To understand the root causes of this generalization gap, we analyze the domain shift from two perspectives: \textit{synthesis-agnostic} degradation from the video processing pipeline, and \textit{model-specific} fingerprints intrinsic to video generation architectures. We conduct a qualitative analysis to examine the potential distribution gap between image and video data. Specifically, we randomly sample data from existing real-world datasets~\cite{arxiv2017kinetics} and AIGC-generated sources~\cite{nips2023genimage, arxiv2023vbench} for both modalities.

\begin{wrapfigure}{r}{0.5\linewidth}
    \centering
    \begin{minipage}{0.48\linewidth}
        \centering
        \includegraphics[width=\textwidth]{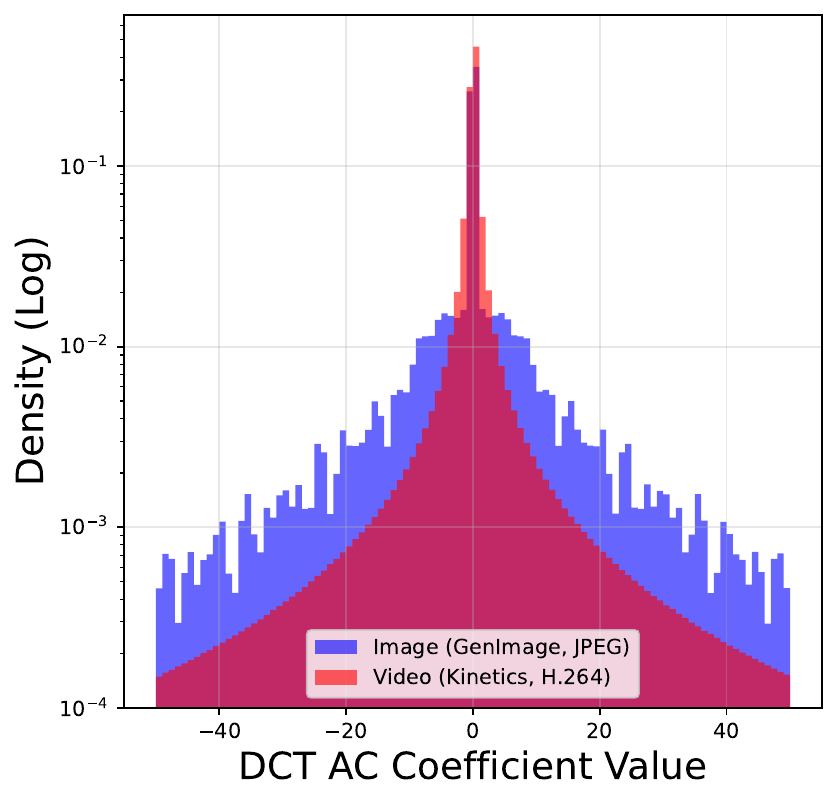}
    \end{minipage}
    \hfill
    \begin{minipage}{0.48\linewidth}
        \centering
        \includegraphics[width=\textwidth]{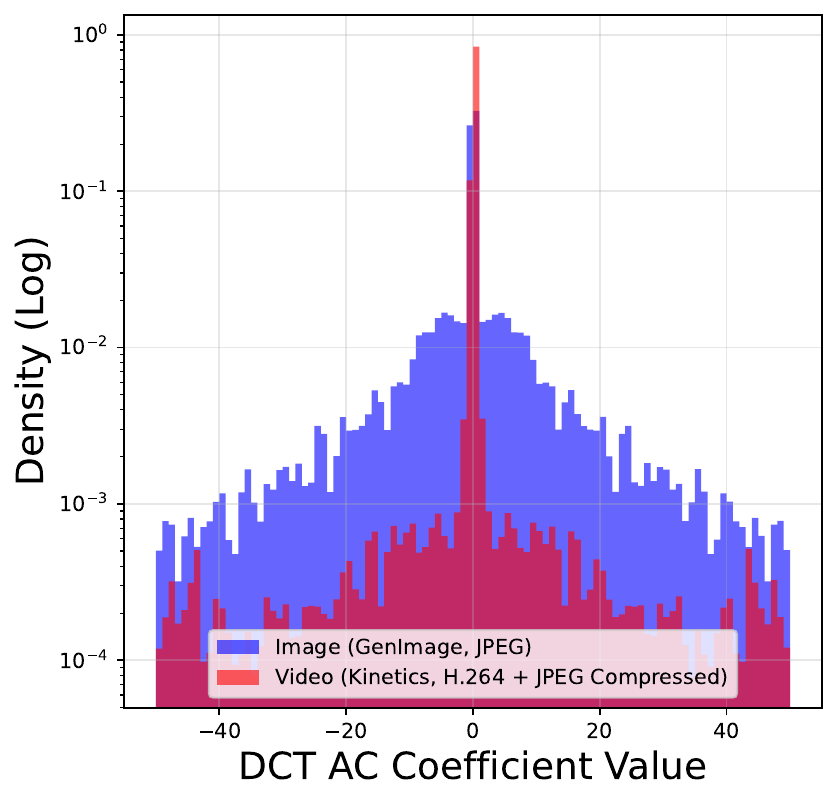}
    \end{minipage}
    \caption{\textbf{DCT AC coefficient distributions.} H.264 and recompressed video frames exhibit sharper near-zero peaks than original JPEG images.}
    \vspace{-6mm}
    \label{fig:analysis_dct}
\end{wrapfigure}

\paragraph{Compression Artifacts.} Quantization noise in videos differs fundamentally from that in still images. As shown in Figure~\ref{fig:analysis_dct}, DCT AC coefficients from video frames display a smoother distribution, with substantially more zeros and a sharper peak near zero compared to JPEG images. This indicates more refined and often more aggressive quantization, driven by inter-frame prediction, variable block sizes, and adaptive quantization. Videos also frequently undergo recompression during distribution, making their DCT distributions even steeper after additional JPEG compression. These distinct frequency-domain statistics can mislead detectors that rely on local pixel cues~\cite{iclr2025sanity} or frequency features~\cite{kdd2025safe}.

\begin{figure}[t]
    \centering
    % \captionsetup{font=small,skip=2pt}
    \begin{minipage}[t]{0.42\linewidth}
        \centering
        \includegraphics[width=0.96\linewidth]{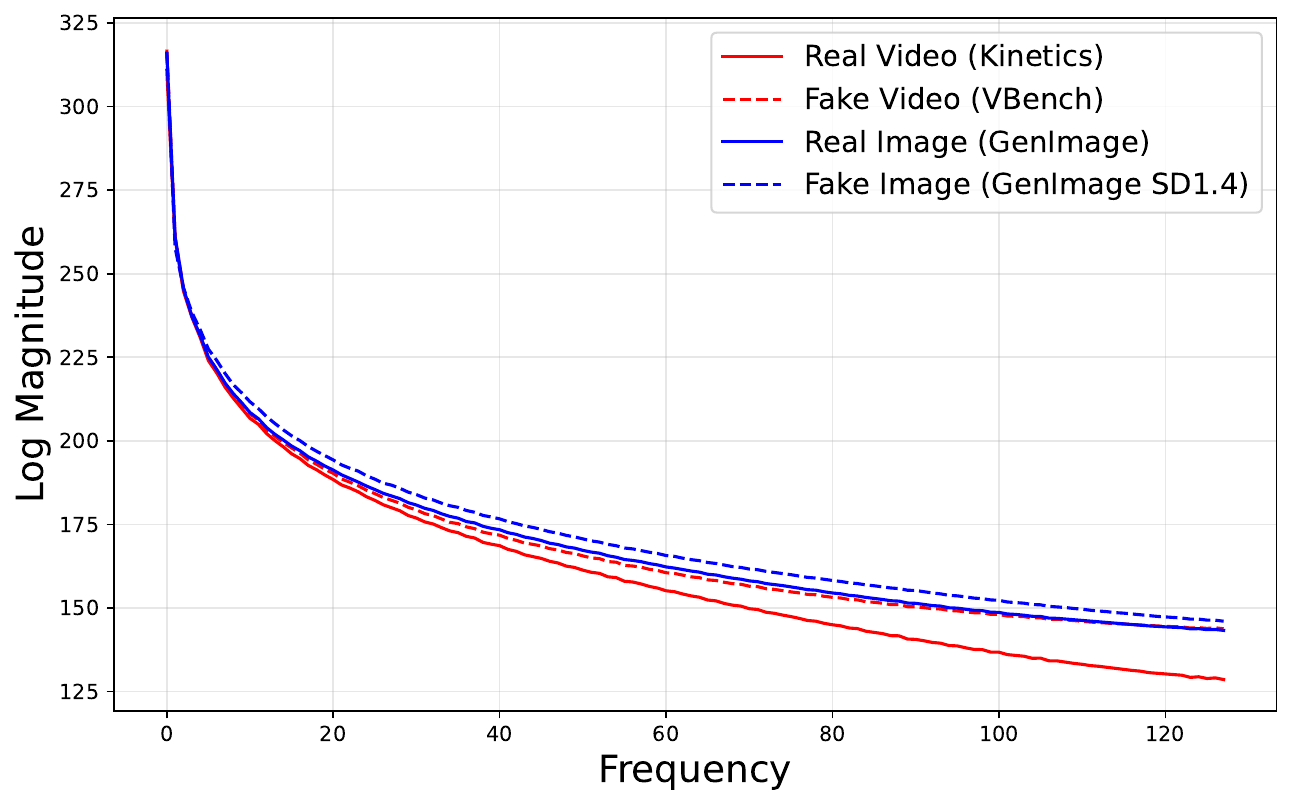}
        \caption{\textbf{RAPSD analysis of video and image datasets.} Real videos exhibit significant high-frequency decay, reflecting compression and motion blur.}
        \label{fig:analysis_psd}
    \end{minipage}
    \hfill
    \begin{minipage}[t]{0.54\linewidth}
        \centering
        \begin{minipage}[t]{0.48\linewidth}
            \centering
            \includegraphics[width=\linewidth]{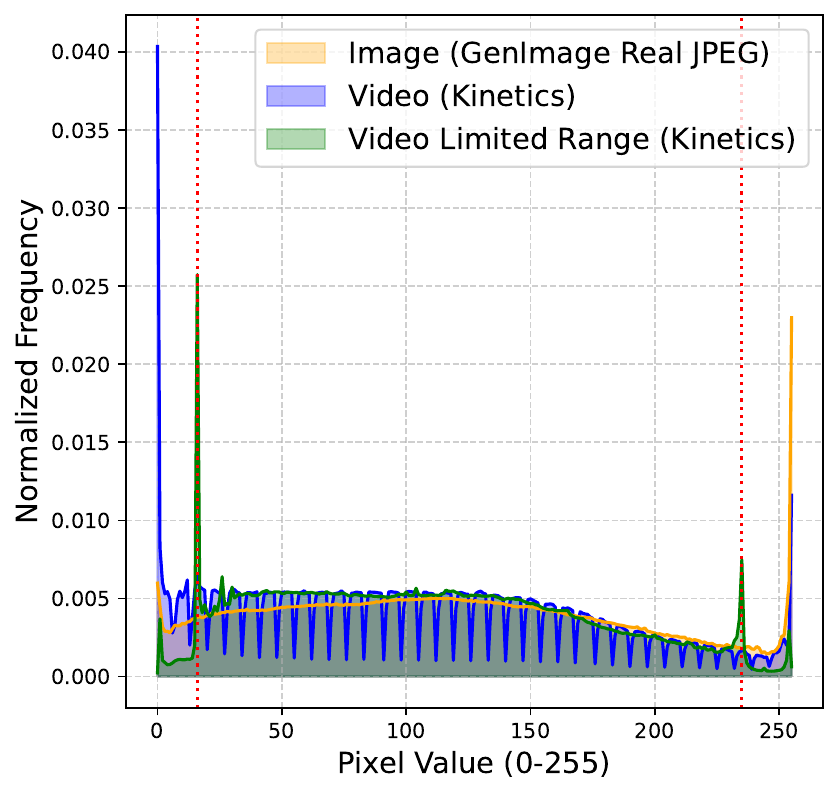}
        \end{minipage}
        \hfill
        \begin{minipage}[t]{0.48\linewidth}
            \centering
            \includegraphics[width=\linewidth]{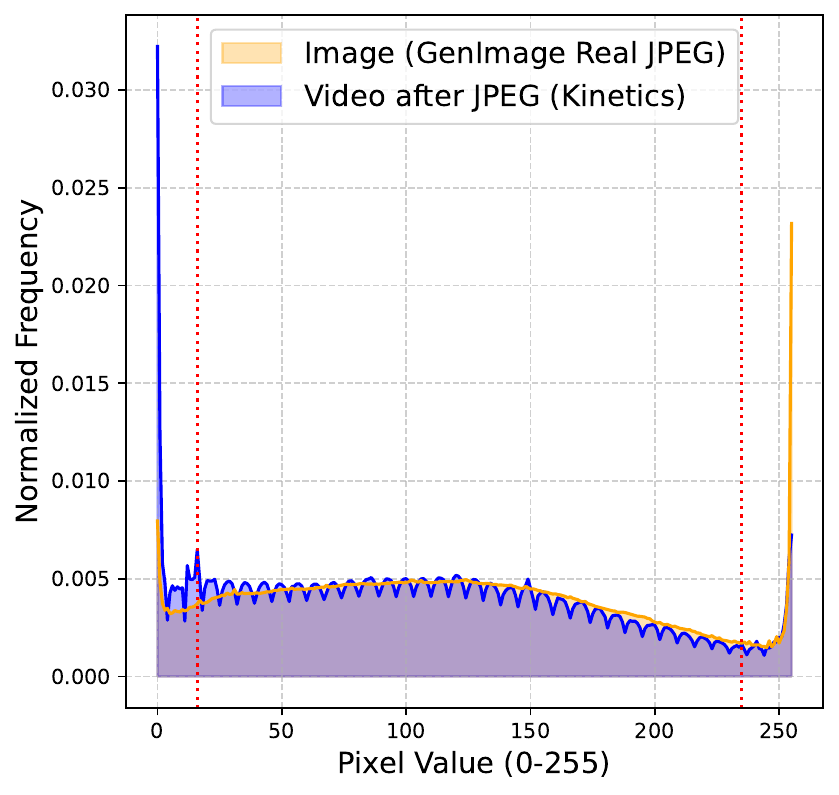}
        \end{minipage}
        \caption{\textbf{Pixel luminance distributions.} We analyze 10k samples each from Kinetics (video) and GenImage (image). Video frames show shifted statistics before and after JPEG double compression.}
        \label{fig:analysis_y_dist}
    \end{minipage}
    \vspace{-2mm}
\end{figure}

\paragraph{Blur and Frequency Decay.} Video frames are inherently prone to spectral degradation due to motion blur, aggressive compression, and resolution rescaling. As illustrated by the Radially Averaged Power Spectral Density (RAPSD)~\cite{cvpr2020watchyourup-convolution} plot in \cref{fig:analysis_psd}, real videos exhibit the most pronounced high-frequency energy decay, suggesting that the video pipeline acts as a strong low-pass filter. The additional visualization in \Cref{sec:appendix_dataset_analysis} (\cref{fig:analysis_dft}) further shows that this filtering substantially reduces the spectral discrepancy between real and synthetic content in the video domain compared to the image domain, helping explain why AIGC detectors degrade under unseen video scenarios.

\paragraph{Color Space and Luminance Range.} \Cref{fig:analysis_y_dist} shows that pixel-intensity statistics shift markedly between images and videos. Most real-world videos use a limited (TV) luminance range (e.g., \(Y \in [16, 235]\)), which removes extreme values and can create empty histogram bins after integer rescaling. While standard augmentation can partially mitigate this discrepancy, detectors designed to capture color mismatches~\cite{cvpr2025secretliesincolor} are prone to fail on such out-of-distribution frames.

\paragraph{Model-Specific Fingerprints.} Beyond pipeline artifacts that differentiate video frames from still images, video generation models introduce distinct, model-driven fingerprints~\cite{nips2025seeingwhatmatters, arxiv2025saga}. Due to temporal modules and specific motion simulation, video generative models often produce unnatural blur and motion-related artifacts, which can manifest as spatial warping or illogical textures in individual frames~\cite{arxiv2025learning}. The spectral visualization in \Cref{fig:analysis_dft} shows that the noise spectrum of AI-generated videos differs noticeably from that of text-to-image-generated images, with additional model-specific patterns provided in \cref{fig:spectrum_model}.

\subsection{Consistency via Video-Image Joint Training}

An ideal AIGC detector should ignore irrelevant pipeline-induced discrepancies and learn robust cues intrinsic to generative models. However, despite alignment efforts~\cite{nips2025dda, nips2025seeingwhatmatters}, intrinsic domain shifts often cause detectors trained solely on images to fail on video frames. At the same time, videos provide rich natural variation that can regularize detectors beyond handcrafted augmentations. To bridge this gap, we propose a joint training paradigm that leverages video data for natural video-augmented consistency learning.

Formally, let $\mathcal{D}_{img}$ be a dataset of static images and $\mathcal{D}_{vid}$ be a dataset of video frames, both annotated with real/fake labels. We model video frames as image signals subjected to a complex, non-differentiable degradation function $\mathcal{T}(\cdot)$. If $x$ represents a pristine visual signal, a video frame $v$ is obtained via:
\begin{equation}
    v = \mathcal{T}(x) = \text{Codec}(\text{Resize}(\text{MotionBlur}(x)))
\end{equation}

Crucially, the transformation $\mathcal{T}(\cdot)$ introduces intricate temporal motion blur and sophisticated compression artifacts. These real-world video degradations are entangled and complex  thus cannot be easily simulated by 2D data augmentations (e.g., Gaussian blur or JPEG re-compression). Consequently, standard detectors trained only on $\mathcal{D}_{img}$ tend to latch onto high-frequency artifacts in $x$ that are inevitably destroyed by the complex dynamics of $\mathcal{T}(\cdot)$. Joint training on image and video samples provides an implicit consistency constraint by exposing the classifier to both pristine and naturally degraded observations. However, relying only on binary supervision may still allow the feature space to remain modality-separated. Inspired by \cite{nips2020supcon}, we add an explicit cross-modal supervised contrastive regularizer that aligns image and video representations with the same real/fake label.

Let $z_i$ denote the $\ell_2$-normalized feature of sample $i$ in a mini-batch, with class label $y_i \in \{0,1\}$ and modality label $m_i \in \{0,1\}$ indicating image or video. For each anchor $i$, we define its positive set using only cross-modal samples with the same class label:
\begin{equation}
\label{eq:cm_supcon_pos_main}
P(i)=\{\,j\neq i:\; y_j=y_i \land m_j\neq m_i\,\}.
\end{equation}
The cross-modal supervised contrastive loss is then
\begin{equation}
\label{eq:cm_supcon_main}
\mathcal{L}_{\mathrm{CM\text{-}SupCon}}
= \frac{1}{|\mathcal{V}|}\sum_{i\in\mathcal{V}}
\frac{-1}{|P(i)|}\sum_{p\in P(i)}
\log
\frac{\exp(z_i^\top z_p/\tau)}
{\sum_{k\neq i}\exp(z_i^\top z_k/\tau)},
\end{equation}
where $\tau$ is the temperature and $\mathcal{V}=\{i:|P(i)|>0\}$ denotes anchors with at least one valid cross-modal positive. This design differs from standard same-label contrastive learning: same-modality samples are not counted as positives, preventing the trivial solution in which image and video features cluster separately while sharing the same class label.

Our final objective combines binary classification with this explicit cross-modal feature constraint:
\begin{equation}
\label{eq:joint_loss}
    \mathcal{L}=\mathcal{L}_{\mathrm{BCE}}+\lambda \mathcal{L}_{\mathrm{CM\text{-}SupCon}}.
\end{equation}
Here, $\mathcal{L}_{\mathrm{BCE}}$ is computed over mixed image and video mini-batches. The CM-SupCon term explicitly pulls real-image features toward real-video features and fake-image features toward fake-video features, while the contrastive denominator preserves real/fake separation. As a result, the detector learns features that are both class-discriminative and less tied to modality-specific compression, blur, or color-range shortcuts.

\section{Experiments}

\subsection{Experimental Setup}

\paragraph{Training Datasets and Evaluation Benchmarks.} We compare three video-image training combinations with different data sources: Pika-100k from GenVideo~\cite{arxiv2024demamba,pika} with ProGAN-4class~\cite{iclr2018progan}, GenVideo-100k~\cite{arxiv2024demamba} with SD1.4 from GenImage~\cite{nips2023genimage}, and the 140k video set from Qwen2.5-ViT~\cite{iclr2026preserving} with DDA~\cite{nips2025dda}. Our evaluation spans 14 diverse benchmarks: four AIGC video detection benchmarks (Magic Videos~\cite{iclr2026preserving}, GenVideo-val~\cite{arxiv2024demamba}, Genbuster++~\cite{arxiv2025busterx++}, DeepTraceReward~\cite{arxiv2025learning}), five AIGC image detection benchmarks (Forensics~\cite{cvpr2020cnngenerated}, UniFD~\cite{cvpr2023unifd}, DiTFake~\cite{kdd2025safe}, ARForensics~\cite{iccv2025d^3qe}), and five in-the-wild benchmarks (Chameleon~\cite{iclr2025sanity}, SynthWildX~\cite{cvprw2024raising}, WildRF~\cite{arxiv2024realtime}, AIGIBench~\cite{nips2025aigibench}, RR-Dataset~\cite{iccv2025rrdataset}). \Cref{tab:dataset-overview} provides a summary. For all benchmarks except GenVideo, we report mean accuracy (ACC) and average precision (AP) across subsets. Due to the severe class imbalance (far fewer fake than real samples) within GenVideo-Val's subsets, we report its overall ACC and AP instead. Full results for each benchmark are provided in \Cref{sec:appendix_benchmark}.

\paragraph{Baselines.} We compare our approach with the following baselines: video-based AIGC detectors: DeMamba~\cite{arxiv2024demamba}, WaveRep~\cite{nips2025seeingwhatmatters}, and Qwen2.5-ViT~\cite{iclr2026preserving}; image-based AIGC detectors: NPR~\cite{cvpr2024npr}, FatFormer~\cite{cvpr2024fatformer}, RINE~\cite{eccv2024rine}, Effort~\cite{icml2025effort}, Co-SPY~\cite{cvpr2025co-spy}, MIRROR~\cite{arxiv2026mirror}; as well as models trained on aligned datasets: B-Free~\cite{cvpr2025b-free}, DDA~\cite{nips2025dda}. For baselines with public checkpoints, we use the officially released checkpoints and preprocessing pipelines; for DeMamba, whose checkpoint is not publicly available, we retrain it on GenVideo-100k. Additional implementation details are provided in \Cref{sec:appendix_implementation}.

\definecolor{videobg}{HTML}{E5F6FF}
\definecolor{imagebg}{HTML}{FFF9E3}
\definecolor{allbg}{HTML}{FDEBFF}

\begin{table*}[tb]
    \centering
    \setlength{\tabcolsep}{2.2pt}
    \renewcommand{\arraystretch}{1.06}
    
    \caption{\textbf{Comparison results on 8 established benchmarks in ACC(\%), including 4 video-based and 4 image-based benchmarks.} Except for DeMamba, for which no public checkpoint is available and which we retrain on GenVideo-100k, all results are obtained by us using the officially released checkpoints and the corresponding inference-time transforms. Detailed results are provided in \Cref{sec:appendix_benchmark}.}
    \label{tab:hybrid_acc}
    
    \resizebox{\linewidth}{!}{%
    \begin{tabular}{@{}l c
    >{\columncolor{videobg}}c
    c c c c
    % @{\hspace{2pt}}
    >{\columncolor{imagebg}}c
    c c c c
    % @{\hspace{2pt}}
    >{\columncolor{allbg}}c@{}}
    
    \toprule
    &
    &
    \multicolumn{5}{>{\columncolor{videobg}}c}{Video}
    &
    \multicolumn{5}{>{\columncolor{imagebg}}c}{Image}
    &
    {ALL}
    \\
    
    \cellcolor{white}Method
    & \cellcolor{white}Type
    & AVG
    & \cellcolor{videobg}Magic
    & \cellcolor{videobg}GenVideo
    & \cellcolor{videobg}\makecell{GenBu\\ster++}
    & \cellcolor{videobg}\makecell{DeepTrace\\Reward}
    & AVG
    & \cellcolor{imagebg}\makecell{ForenSynths\\(GAN)}
    & \cellcolor{imagebg}\makecell{UniFD\\(LDM)}
    & \cellcolor{imagebg}DiTFake
    & \cellcolor{imagebg}\makecell{ARForen\\sics}
    & AVG
    \\
    \midrule
    DeMamba~\cite{arxiv2024demamba} (100k) & Video & 69.2 & 63.1 & 90.4 & 51.5 & 71.9 & 57.7 & 54.6 & 57.4 & 63.2 & 55.8 & 63.5 \\
    Qwen2.5-ViT~\cite{iclr2026preserving} & Video & \underline{85.0} & 81.3 & 95.6 & 68.8 & 94.4 & 65.4 & 51.4 & 60.0 & 80.3 & 69.9 & 75.2  \\
    WaveRep~\cite{nips2025seeingwhatmatters} & Video & 72.4 & 56.6 & 91.5 & 53.4 & 88.3 & 81.1 & 78.6 & 80.5 & 86.4 & 79.1 & 76.8 \\
    \midrule
    NPR~\cite{cvpr2024npr} & Image & 51.7 & 50.1 & 55.4 & 50.9 & 50.5 & 80.3 & 76.2 & 92.8 & 76.9 & 75.3 & 66.0 \\
    FatFormer~\cite{cvpr2024fatformer} & Image & 52.4  & 49.5 & 56.5 & 51.4  & 52.0  & 80.4  & 88.9  & 93.5  & 60.0  & 79.1  & 66.4  \\
    RINE~\cite{eccv2024rine} & Image & 51.5  & 42.2 & 56.5 & 58.4 & 48.9 & 79.6  & 91.8 & 90.5 & 57.0 & 79.0 & 65.5 \\
    Effort~\cite{icml2025effort} & Image & 69.5 & 67.4 & 79.5 & 62.8 & 68.3 & 82.3 & 73.5 & 91.7 & 82.2 & 81.7 & 75.9 \\
    Co-Spy~\cite{cvpr2025co-spy} & Image & 72.4 & 65.0 & 79.9 & 69.2 & 75.3 & 75.4 & 62.2 & 76.4 & 93.4 & 69.4 & 73.9 \\
    Co-Spy (ProGAN) & Image & 50.9 & 49.7 & 54.1 & 50.1 & 50.0 & 88.4 & 83.0 & 94.4 & 90.8 & 85.6 & 69.7 \\
    DDA~\cite{nips2025dda} & Image & 68.9 & 73.9 & 63.2 & 65.5 & 73.1 & 87.1 & 81.7 & 80.6 & 98.6 & 87.5 & 78.0 \\
    B-Free~\cite{cvpr2025b-free} & Image & 70.6 & 71.6  & 80.8 & 58.3  & 71.9 & 91.2 & 88.2 & 89.1 & 95.5  & 91.9  & 80.9   \\
    \midrule
    VINA (Pika+ProGAN) & V+I & 79.5 & 64.3 & 90.8 & 80.1 & 82.9 & \underline{92.0} & 82.0 & 94.8 & 97.2 & 94.1 & 85.8 \\
    VINA (GenVideo+SD1.4) & V+I & 82.9 & 72.0 & 95.7 & 71.2 & 92.7 & 91.8 & 88.9 & 94.5 & 95.1 & 88.5 & \underline{87.4} \\
    % VINA (140k+DDA) & V+I & \textbf{90.3} & 80.2 & 96.8 & 86.4 & 97.9 & \textbf{93.0} & 85.7 & 94.4 & 95.8 & 96.0 & \textbf{91.6} \\
    VINA (140k+DDA) & V+I & \textbf{90.4} & 77.1 & 96.9 & 90.1 & 97.4 & \textbf{94.1} & 87.0 & 94.8 & 98.0 & 96.7 & \textbf{92.3} \\
    \bottomrule
    \end{tabular}
    \vspace{-1em}
    }
\end{table*}

\paragraph{Implementation Details.} Unless otherwise specified, we use DINOv3-Large as the visual backbone and attach a single MLP classification layer, with all parameters tuned end-to-end. Images and video frames are processed at a fixed resolution of $256\times256$. 
%Video data are first decoded into JPEG frames at 2 FPS to better match the storage format of image inputs. 
To handle diverse aspect ratios, each input is resized with its original aspect ratio preserved until the shorter side reaches 256 pixels, followed by a $256\times256$ crop. During training, we sample one frame per video clip, apply random frame sampling and random cropping, and use only random horizontal flipping as additional augmentation. During inference, we select the middle frame of each video and apply center cropping. We deliberately avoid extra data augmentation or JPEG alignment to isolate the effect of video-image joint training. Models are trained with AdamW for 10 epochs using a fixed learning rate of $1\times10^{-6}$, a batch size of 64, and early stopping based on validation loss. The supervised contrastive term uses $\lambda=0.05$ in the final objective, with a temperature $\tau=0.07$ in the CM-SupCon loss. All reported results are produced by a single model without dataset-specific parameter tuning.

\subsection{Cross-Dataset Evaluation}

\paragraph{Compared to Video-based and Image-based Detectors.} As indicated in \cref{tab:hybrid_acc}, \emph{video-based detectors} exhibit asymmetric generalization. Qwen2.5-ViT~\cite{iclr2026preserving} performs strongly on video benchmarks but degrades on images, while DeMamba~\cite{arxiv2024demamba} shows limited transfer across both modalities when retrained on GenVideo-100k. WaveRep~\cite{nips2025seeingwhatmatters} fine-tunes a DINOv2 image backbone on video data, preserving better image generalization but yielding weaker video performance. In contrast, our jointly trained detector reaches 90.4\% average ACC on video benchmarks using only one frame at inference. Moreover, \emph{image-based detectors} show a pronounced cross-modal failure. Early methods such as NPR~\cite{cvpr2024npr}, FatFormer~\cite{cvpr2024fatformer}, and RINE~\cite{eccv2024rine} perform well on several image benchmarks but often collapse on videos. More recent detectors trained with aligned datasets, such as DDA~\cite{nips2025dda} and B-Free~\cite{cvpr2025b-free}, improve cross-dataset generalization but still suffer substantial video-domain degradation. Our method consistently outperforms these image-only detectors on both image and video benchmarks, demonstrating stronger cross-modal robustness.

\paragraph{Joint Video-Image Training.} To isolate the effect of video data, we deliberately avoid complex augmentation and high-resolution processing widely applied in image-based methods~\cite{nips2025dda,cvpr2025b-free}; nevertheless, the proposed strategy already achieves state-of-the-art performance. Our method uses video frames as natural augmentations and consistency regularizers, and its gains are not tied to a specific source: both early combinations such as Pika+ProGAN and higher-quality mixtures such as 140k+DDA substantially outperform single-modality baselines, with more detailed evidence in \Cref{tab:ablation_i+v_right}. Moreover, our approach is complementary to explicit alignment, augmentation, and post-processing techniques, which can be combined with joint training to further raise the performance ceiling.

\newcolumntype{Y}{>{\centering\arraybackslash}X}
\renewcommand{\tabularxcolumn}[1]{m{#1}}

\begin{table}[t]
    \caption{\textbf{Benchmarking results on in-the-wild benchmarks in terms of balanced ACC(\%).} $\dagger$ indicates that the results are obtained from their papers.}
    \label{tab:in_the_wild}
    \centering
    % \resizebox{\linewidth}{!}{
    \begin{adjustbox}{max width=\textwidth}
    \begin{tabular}{l|c|ccc|ccc|cc|c|c}
    \toprule
    \multirow{2}{*}{Method} & \multirow{2}{*}{\makecell{Chame\\-leon~\cite{iclr2025sanity}}} &  \multicolumn{3}{c|}{SynthWildx~\cite{cvprw2024raising}} & \multicolumn{3}{c|}{WildRF~\cite{arxiv2024realtime}} & \multicolumn{2}{c|}{AIGIBench~\cite{nips2025aigibench}} & \multirow{2}{*}{\makecell{RR-Da\\taset~\cite{iccv2025rrdataset}}} & \multirow{2}{*}{AVG} \\
    & & DALLE3 & Firefly & Midj. & FB & Reddit & Twitter & SocRF & ComAI & & \\
    \midrule
    WaveRep~\cite{nips2025seeingwhatmatters} & 67.9 & 72.9 & 94.9 & 70.2 & 75.3 & 77.3 & 82.9 & 68.6 & 72.2 & 65.1 & 74.7 \\
    % AIDE$\dagger$~\cite{iclr2025sanity} & 65.8 \\
    NPR~\cite{cvpr2024npr} & 53.1 & 40.6 & 63.5 & 46.5 & 74.1 & 63.8 & 48.5 & 53.3 & 55.1 & 60.9 & 55.9 \\
    FatFormer~\cite{cvpr2024fatformer} & 51.1 & 47.4 & 59.7 & 49.8 & 54.4 & 69.5 & 42.5 & 56.9 & 51.9 & 53.0 & 53.8 \\
    RINE~\cite{eccv2024rine} & 70.0 & 45.7 & 68.6 & 48.0 & 52.5 & 67.8 & 45.8 & 56.0 & 51.2 & 52.1 & 55.8 \\
    Effort~\cite{icml2025effort} & 65.4 & 51.6 & 55.7 & 47.9 & 57.8 & 58.3 & 67.9 & 51.4 & 56.8 & 56.5 & 56.9 \\
    CO-SPY~\cite{cvpr2025co-spy} & 78.4 & 72.0 & 79.7 & 70.8 & 73.1 & 77.3 & 75.3 & 68.2 & 66.7 & 71.0 & 73.3 \\
    DDA~\cite{nips2025dda} & 82.4 & 92.4 & 87.2 & 93.1 & 93.1 & 86.4 & 91.2 & 81.7 & 84.6 & 76.0 & 86.8 \\
    B-Free~\cite{cvpr2025b-free} & 76.6 & 95.7  & 95.8 & 95.6  & 95.3  & 86.5  & 97.9  & 85.2  & 81.5  & 72.3  & 88.2 \\
    MIRROR$\dagger$~\cite{arxiv2026mirror} & 90.7 & 95.9 & 88.4 & 94.9 & 97.1 & 96.6 & 96.4 & 87.6 & 93.4 & 78.9 & \underline{92.0} \\
    \midrule
    VINA (Pika+ProGAN) & 86.8 & 75.9 & 76.6 & 75.7 & 87.2 & 83.7 & 91.2 & 74.2 & 89.9 & 67.4 & 80.9  \\
    VINA (GenVideo+SD1.4) & 88.7 & 93.1 & 89.2 & 94.1 & 96.9 & 95.5 & 96.7 & 90.9 & 91.5 & 77.5 & 91.4 \\
    VINA (140k+DDA) & 91.4 & 94.5 & 89.2 & 94.7 & 96.6 & 96.7 & 98.1 & 90.9 & 93.9 & 82.7 & \textbf{92.9} \\
    \bottomrule
    \end{tabular}
    \end{adjustbox}
    \vspace{-2mm}
    % }
\end{table}

\subsection{Comparison on In-the-Wild Benchmarks}
In-the-wild benchmarks better reflect real deployment scenarios, where samples may come from unknown generators, undergo unknown post-processing, and be further compressed or degraded by online platforms. As shown in \Cref{tab:in_the_wild}, early image detectors such as NPR~\cite{cvpr2024npr}, FatFormer~\cite{cvpr2024fatformer}, and RINE~\cite{eccv2024rine} largely fail under these conditions, indicating limited generalization and strong reliance on shortcut cues. Methods trained with aligned datasets or strong augmentations, such as DDA~\cite{nips2025dda}, B-Free~\cite{cvpr2025b-free}, and WaveRep~\cite{nips2025seeingwhatmatters}, achieve better overall robustness, but still show clear weaknesses on challenging benchmarks such as Chameleon and RR-Dataset. Our method remains competitive with and often outperforms these methods, including MIRROR~\cite{arxiv2026mirror}, which also uses a DINOv3 backbone. These results show that joint video-image training provides strong in-the-wild generalization without relying on a specific data source or complex alignment and augmentation pipelines, underscoring both the necessity of unified AIGC detection and the effectiveness of our approach.

\subsection{Ablation Study}

We conduct ablations to examine three factors behind the proposed joint training strategy: whether the gain is consistent across vision backbones, whether it depends on particular image or video data sources, and whether synthetic degradation strategies can bridge the same video-image gap.

\begin{table}[t]
    \begin{minipage}[t]{0.45\textwidth}
    \centering
    \captionof{table}{\textbf{Ablation Study (ACC) on Joint Video--Image Training using Different Vision Backbones.} 
    I=Image Dataset (GenImage SD1.4), V=Video Dataset (140k).}
    \label{tab:ablation_i+v_left}
    \centering
    \resizebox{\linewidth}{!}{
    \begin{tabular}{cccccc}
    \toprule
    Backbone & \makecell{Training\\Data} & \makecell{Video\\AVG} & \makecell{Image\\AVG} & \makecell{Chame\\leon} & \makecell{Mean\\ACC} \\
    \midrule
    \multirow{3}{*}{\makecell{CLIP ViT-L\\~\cite{arxiv2021clip}}}
    & I & 51.6 & 70.5 & 57.9 & 60.0 \\
    & V & 87.4 & 63.3 & 43.8 & 64.8 \\
    & V+I & \textbf{88.0} & \textbf{84.8} & \textbf{84.7} & \textbf{85.7} \\
    \midrule
    \multirow{3}{*}{\makecell{DINOv2 ViT-L\\~\cite{oquab2023dinov2}}}
    & I & 59.5 & 66.8 & 58.2 & 61.5 \\
    & V & 88.8 & 70.2 & 46.6 & 68.6 \\
    & V+I & \textbf{89.2} & \textbf{86.3} & \textbf{84.0} & \textbf{86.5} \\
    \midrule
    \multirow{3}{*}{\makecell{DINOv3 ViT-L\\~\cite{arxiv2025dinov3}}}
    & I & 59.6 & 75.1 & 59.5 & 64.7 \\
    & V & 88.7 & 79.8 & 49.5 & 72.7 \\
    & V+I & \textbf{89.1} & \textbf{88.4} & \textbf{89.0} & \textbf{88.8} \\
    \bottomrule
    \end{tabular}
    }
    \end{minipage}
    \begin{minipage}[t]{0.54\textwidth}
    \centering
    \captionof{table}{\textbf{Ablation Study on Joint Video--Image Training with Different Data Sources (ACC).} Backbone is DINOv3-L.}
    \label{tab:ablation_i+v_right}
    \centering
    \resizebox{\linewidth}{!}{
    \begin{tabular}{lccccc}
    \toprule
    Training Data & Num & \makecell{Video\\AVG} & \makecell{Image\\AVG} & \makecell{Chame\\leon} & \makecell{Mean\\ACC} \\
    \midrule
    DDA (Image) & 236k & 68.4 & 89.8 & 78.7 & 79.0 \\
    DDA+SD1.4 & 560k & 67.8 & 86.9 & 65.0 & 73.2\\
    DDA+SD1.4+ProGAN & 705k & 64.7 & 88.8 & 76.8 & 76.8 \\
    \midrule
    140k (Video) & 140k & 88.7 & 79.8 & 49.5 & 72.6 \\
    140k+GenVideo & 340k & 85.8 & 77.5 & 46.7 & 70.0 \\
    \midrule
    GenVideo+SD1.4 & 524k & 82.9 & 91.8 & 89.4 & 88.0 \\
    GenVideo+DDA & 436k & 80.2 & 91.8 & 86.8 & 86.3 \\
    % 140k+ProGAN & \\
    140k+SD1.4 & 464k & 89.1 & 88.4 & 89.0 & 88.8 \\
    140k+DDA & 376k & \textbf{90.4} & \textbf{94.1} & \textbf{91.4} & \textbf{92.0} \\
    \bottomrule
    \end{tabular}
    }
    \end{minipage}
\end{table} 

\paragraph{Ablation Study on Model Backbone.} As shown in \Cref{tab:ablation_i+v_left}, joint image-video (I+V) training is consistently effective across CLIP, DINOv2, and DINOv3 backbones. More importantly, the gains are bidirectional. Image-only training performs poorly on videos, while video-only training improves video accuracy but substantially hurts image and Chameleon performance. Combining both modalities resolves this trade-off: adding video data greatly improves video robustness, and adding image data restores strong image and in-the-wild generalization. This demonstrates that the benefit comes from complementary cross-modal supervision rather than a particular backbone.

\paragraph{Ablation Study on Data Sources.} \Cref{tab:ablation_i+v_right} further shows that the proposed strategy is not tied to a specific data source. Different image-video combinations all achieve strong performance, whereas simply scaling up image-only data (DDA+SD1.4 or DDA+SD1.4+ProGAN) or video-only data (140k+GenVideo) does not improve generalization and can even degrade the mean accuracy. In contrast, joint training achieves substantially stronger results with smaller data mixtures, indicating that the gain comes from the joint image-video training itself rather than from increased data scale or carefully selected sources.

\begin{table}[t]
  \begin{minipage}[t]{0.67\textwidth}

    \caption{\textbf{Ablations on Strategies for Bridging the Image-Video Gap}. We compare data augmentations designed for format adaptation with our proposed Joint training approach.}
    \label{tab:ablation_format}
    \centering
    \resizebox{\linewidth}{!}{
    \begin{tabular}{l|cccc}
    \toprule
    Method & \makecell{Video\\AVG} & \makecell{Image\\AVG} & \makecell{Chame\\leon} & \makecell{Mean\\ACC} \\
    \midrule
    % ProGAN & I & \\
    DINOv3 (trained on DDA) & 68.4 & 89.8 & 78.7 & 78.9 \\
    % random JPEG compression &  \\
     + random color jitter & 68.5 & 91.0 & 81.3 & 80.3 \\
     + random chain degradation & 71.5 & 88.8 & 85.4 & 81.9 \\
     + random HEIF compression & 73.5 & 88.9 & 85.5 & 82.6 \\
     + random WebP compression & 72.4 & 89.8 & 88.9 & 83.7 \\
    DINOv3 (DDA+140k) & 89.8 & 92.9 & 90.4 & \underline{90.9} \\
     + CM-SupCon loss (Ours) & 90.4 & 94.1 & 91.4 & \textbf{92.0} \\
    \bottomrule
    \end{tabular}
    }
    \end{minipage}
    \hfill
  \begin{minipage}[t]{0.31\textwidth}

    \caption{\textbf{Image Performance Comparison after JPEG Compression (Q=96).}}
    \label{tab:jpeg_alignment}
    \centering
    \resizebox{\linewidth}{!}{
    \begin{tabular}{cc|cccc}
    \toprule
    Method & JPEG & \makecell{Image\\ACC} & \makecell{Image\\AP} \\
    \midrule
    % ProGAN & I & \\
    SAFE & & 94.8 & 96.6 \\
    ~\cite{kdd2025safe} & \checkmark & 50.2 & 52.0 \\
    \midrule
    CO-SPY & & 88.4 & 94.0 \\
    ~\cite{cvpr2025co-spy} & \checkmark & 53.1 & 54.6 \\
    \midrule
    Effort & & 82.3 & 89.1 \\
    ~\cite{icml2025effort} & \checkmark & 67.5 & 77.0 \\
    \midrule
    \multirow{2}{*}{VINA} & & 87.3 & 97.2 \\
     & \checkmark & 86.3 & 97.1 \\
    \bottomrule
    \end{tabular}
    }
  \end{minipage}
  \vspace{-1em}
\end{table}

\paragraph{Ablation Study on Video-Image Gap Bridge Strategy.} Motivated by the analysis in \cref{sec:method_analysis}, we compare several synthetic degradation strategies that approximate video-domain shifts, including color jitter, HEIF/WebP compression, and a random degradation chain. As shown in \Cref{tab:ablation_format}, these augmentations moderately improve video and in-the-wild performance, confirming that compression and color-space shifts contribute to the image-video gap. However, their gains remain much smaller than direct I+V training, suggesting that real video data provides richer and harder-to-simulate degradations. Adding the supervised contrastive loss further improves performance by aligning image and video representations, yielding the best overall result.

\subsection{Analysis}

\begin{figure}[tbp]
  \begin{minipage}[t]{0.98\textwidth}
    \centering
    \includegraphics[width=\linewidth]{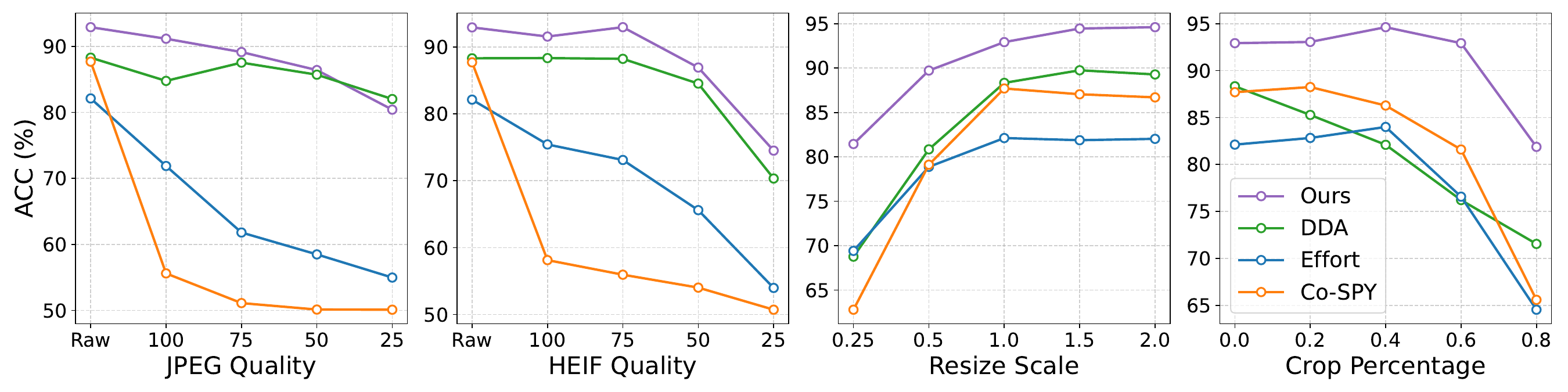}
  \end{minipage}
    \caption{\textbf{Robustness Analysis Across Various Perturbations on Established Image Benchmarks.} Degradations include JPEG compression, HEIF compression, scaling, and cropping. Our method consistently outperforms other detectors, demonstrating superior robustness.}
    \label{fig:robustness}
    \vspace{-1em}
\end{figure}

\paragraph{Robustness Analysis.} As observed in prior work~\cite{iclr2025aligned,nips2025dda}, many AIGC image datasets suffer from a \emph{format bias}: real images are typically JPEG-compressed, while generated images are often stored as PNGs. We align formats by applying JPEG compression ($Q=96$) to all fake PNG images in our four image evaluation sets. As shown in \Cref{tab:jpeg_alignment}, this causes a substantial drop for most detectors, whereas our model trained on 140k+SD1.4 remains stable without explicit image-level alignment. The broader perturbation results in \Cref{fig:robustness} further show consistent robustness across JPEG compression, HEIF compression, scaling, and cropping, suggesting that joint video-image training suppresses shortcut learning through natural video regularization.

\begin{table}[t]
    \begin{minipage}[t]{0.49\textwidth}
    \centering
    \captionof{table}{\textbf{Computational Efficiency Comparison.} One RTX 5090 GPU is used in inference.}
    \label{tab:efficiency}
    \resizebox{\linewidth}{!}{
    \begin{tabular}{lcccc}
    \toprule
    Method & Params & Resolution & FLOPs & \makecell{Time\\ / 1K Frames} \\
    \midrule
    DDA & 306.7M & $336\times336$ & 176.6G & 22.3s \\
    B-Free & 86.5M & $504\times504$ & 553.7G & 77.4s \\
    WaveRep & 86.5M & $504\times504$ & 111.2G & 24.1s \\
    CO-SPY & 513.3M & $384\times384$ & 645.7G & 61.8s \\
    Ours & 303.1M & $256\times256$ & \textbf{82.4G} & \textbf{15.7s} \\
    \bottomrule
    \end{tabular}
    }
    \end{minipage}
    \hfill
    \begin{minipage}[t]{0.49\textwidth}
    \centering
    \captionof{table}{\textbf{Resolution and Backbone Scaling.} We train DINOv3 models with scaled resolution or backbone size.}
    \label{tab:scaling_spatial}
    \resizebox{\linewidth}{!}{
    \begin{tabular}{cc|cccc}
    \toprule
    Backbone & Resolution & Video & Image & Chameleon & Mean \\
    \midrule
    \multirow{4}*{ViT-L 300M} & $256^2$ & 90.4 & 94.1 & 91.4 & 92.0 \\
    & $384^2$ & 89.4 & 96.6 & 91.6 & 92.5 \\
    & $448^2$ & 90.0 & 96.2 & 91.6 & 92.6 \\
    & $512^2$ & 92.5 & 98.4 & 91.0 & \textbf{94.0} \\
    \midrule
    ViT-H+ 840M & $256^2$ & 90.1 & 95.9 & 94.0 & \underline{93.3} \\
    \bottomrule
    \end{tabular}
    }
    \end{minipage}
    \vspace{-1em}
\end{table}

\paragraph{Computational Efficiency.} \Cref{tab:efficiency} compares the computational cost of representative image and video AIGC detectors under a unified measurement protocol. All methods are evaluated on a single NVIDIA RTX 5090 GPU using FP32 precision, and the batch size is set to 64 or the maximum value allowed by GPU memory for each method. Despite using a large DINOv3 backbone, our model processes frames at $256\times256$ resolution and requires only 82.4G FLOPs per frame, yielding the fastest throughput at 15.7s per 1K frames. This indicates that the proposed unified detector is not only accurate across image and video benchmarks but also computationally efficient at inference time.

\paragraph{Scaling Spatial Resolution and Backbone.} \Cref{tab:scaling_spatial} evaluates spatial scalability by retraining models at higher input resolutions and with larger backbones. For ViT-L (300M), increasing resolution from $256^2$ to $512^2$ yields a monotonic improvement in mean accuracy (92.0 $\rightarrow$ \textbf{94.0}), with consistent gains on both image and video benchmarks. Scaling the backbone to ViT-H+ (840M) further improves the $256^2$ setting, suggesting that the framework benefits from both higher-resolution inputs and increased model capacity.

\FloatBarrier
\section{Conclusion}

This paper investigates why state-of-the-art AI-generated image detectors often fail to generalize to video frames.
Through an analysis of real-world video acquisition and distribution, we identify synthesis-agnostic domain shifts that overwhelm image-centric forensic cues and model-specific artifacts introduced by modern video generators.
Guided by these insights, we propose VINA, a unified AIGC detection framework and training philosophy that treats realistic video degradations as physically grounded augmentation and enforces consistency across high-quality images and degraded video-style frames.
Extensive experiments show that VINA substantially improves cross-domain robustness and achieves strong performance in challenging in-the-wild scenarios.
\paragraph{Limitations.} Our detector operates primarily on single frames and does not explicitly model temporal relationships, which may miss artifacts that only emerge through motion inconsistency. In addition, current analysis of model interpretability remains limited. Incorporating temporal reasoning and more faithful attribution analysis could further improve robustness, accuracy, and transparency.

\bibliography{abrv, reference}

@String(CVPR= {IEEE Conf. Comput. Vis. Pattern Recog.})

@String(ICCV= {Int. Conf. Comput. Vis.})

@String(ECCV= {Eur. Conf. Comput. Vis.})

@String(NIPS= {Adv. Neural Inform. Process. Syst.})

@String(ICLR = {Int. Conf. Learn. Represent.})

@String(AAAI = {AAAI})

@String(CVPRW= {IEEE Conf. Comput. Vis. Pattern Recog. Worksh.})

@String(CVPR  = {CVPR})

@String(ICCV  = {ICCV})

@String(ECCV  = {ECCV})

@String(NIPS  = {NeurIPS})

@String(ICML  = {ICML})

@String(ICLR  = {ICLR})

@String(CVPRW= {CVPRW})

@misc{sora,
    author = {Tim Brooks and Bill Peebles and Connor Holmes and Will DePue and Yufei Guo and Li Jing and David Schnurr and Joe Taylor and Troy Luhman and Eric Luhman and Clarence Ng and Ricky Wang and Aditya Ramesh},
    year = {2024},
    url = {https://openai.com/index/sora/},
    title = {Video generation models as world simulators}
}

@misc{kling,
    author = {Kuaishou},
    year = {2024},
    url = {https://klingai.kuaishou.com},
    title = {https://klingai.kuaishou.com}
}

@misc{pika,
    author = {{Pika Labs}},
    year = {2023},
    url = {https://pika.art/},
    title = {https://pika.art/}
}

@article{wang2024aigcsecurity,
  title={Security and privacy on generative data in aigc: A survey},
  author={Wang, Tao and Zhang, Yushu and Qi, Shuren and Zhao, Ruoyu and Xia, Zhihua and Weng, Jian},
  journal={ACM Computing Surveys},
  volume={57},
  number={4},
  pages={1--34},
  year={2024},
  publisher={ACM New York, NY}
}

@inproceedings{
    iclr2026preserving,
    title={Preserving Forgery Artifacts: {AI}-Generated Video Detection at Native Scale},
    author={Zhengcen Li and Chenyang Jiang and Hang Zhao and Shiyang Zhou and Yunyang Mo and Feng Gao and Fan Yang and Qiben Shan and Shaocong Wu and Jingyong Su},
    booktitle=ICLR,
    year={2026},
    url={https://openreview.net/forum?id=XD43lfRCg6}
}

@misc{arxiv2026aligngemini,
  title = {AlignGemini: Generalizable AI-Generated Image Detection Through Task-Model Alignment},
  author = {Chen, Ruoxin and Gao, Jiahui and Lin, Kaiqing and Zhang, Keyue and Zhao, Yandan and Guan, Isabel and Yao, Taiping and Ding, Shouhong},
  year = 2026,
  number = {arXiv:2512.06746},
  eprint = {2512.06746},
  primaryclass = {cs},
  archiveprefix = {arXiv}
}

@misc{arxiv2026videoveritas,
  title = {VideoVeritas: AI-Generated Video Detection via Perception Pretext Reinforcement Learning},
  author = {Tan, Hao and Lan, Jun and Shi, Senyuan and Tan, Zichang and Yu, Zijian and Zhu, Huijia and Wang, Weiqiang and Wan, Jun and Lei, Zhen},
  year = 2026,
  number = {arXiv:2602.08828},
  eprint = {2602.08828},
  primaryclass = {cs},
  archiveprefix = {arXiv},
  note = {arXiv:2602.08828},
}

@inproceedings{nips2020supcon,
  title = {Supervised Contrastive Learning},
  booktitle = NIPS,
  author = {Khosla, Prannay and Teterwak, Piotr and Wang, Chen and Sarna, Aaron and Tian, Yonglong and Isola, Phillip and Maschinot, Aaron and Liu, Ce and Krishnan, Dilip},
  year = 2020,
  volume = {33},
  pages = {18661--18673}
}

@misc{arxiv2025beyondartifacts,
  title = {Beyond Artifacts: Real-Centric Envelope Modeling for Reliable AI-Generated Image Detection},
  author = {Liu, Ruiqi and Han, Yi and Zhang, Zhengbo and Yao, Liwei and Yan, Zhiyuan and Shen, Jialiang and Chen, ZhiJin and Sun, Boyi and Weng, Lubin and Dong, Jing and Wang, Yan and Wu, Shu},
  year = 2025,
  number = {arXiv:2512.20937},
  eprint = {2512.20937},
  primaryclass = {cs},
  archiveprefix = {arXiv},
  note = {arXiv:2512.20937},
}

@inproceedings{iccv2025rrdataset,
  title = {Bridging the Gap Between Ideal and Real-World Evaluation: Benchmarking AI-Generated Image Detection in Challenging Scenarios},
  booktitle = ICCV,
  author = {Li, Chunxiao and Wang, Xiaoxiao and Li, Meiling and Miao, Boming and Sun, Peng and Zhang, Yunjian and Ji, Xiangyang and Zhu, Yao},
  year = 2025,
  eprint = {2509.09172},
  primaryclass = {cs},
  archiveprefix = {arXiv}
}

@inproceedings{nips2025aigibench,
  title = {Is Artificial Intelligence Generated Image Detection a Solved Problem?},
  booktitle = NIPS,
  author = {Li, Ziqiang and Yan, Jiazhen and He, Ziwen and Zeng, Kai and Jiang, Weiwei and Xiong, Lizhi and Fu, Zhangjie},
  year = 2025,
  eprint = {2505.12335},
  primaryclass = {cs},
  archiveprefix = {arXiv}
}

@misc{arxiv2024realtime,
  title = {Real-Time Deepfake Detection in the Real-World},
  author = {Cavia, Bar and Horwitz, Eliahu and Reiss, Tal and Hoshen, Yedid},
  year = 2024,
  number = {arXiv:2406.09398},
  eprint = {2406.09398},
  primaryclass = {cs},
  archiveprefix = {arXiv},
  note = {arXiv:2406.09398},
}

@misc{arxiv2026mirror,
  title = {MIRROR: Manifold Ideal Reference ReconstructOR for Generalizable AI-Generated Image Detection},
  author = {Liu, Ruiqi and Cui, Manni and Qin, Ziheng and Yan, Zhiyuan and Chen, Ruoxin and Han, Yi and Li, Zhiheng and Chen, Junkai and Chen, ZhiJin and Lin, Kaiqing and Shen, Jialiang and Weng, Lubin and Dong, Jing and Wang, Yan and Wu, Shu},
  year = 2026,
  number = {arXiv:2602.02222},
  eprint = {2602.02222},
  primaryclass = {cs},
  archiveprefix = {arXiv},
  note = {arXiv:2602.02222},
}

@inproceedings{aaai2025c2p-clip,
  title = {C2P-CLIP: Injecting Category Common Prompt in CLIP to Enhance Generalization in Deepfake Detection},
  booktitle = AAAI,
  author = {Tan, Chuangchuang and Tao, Renshuai and Liu, Huan and Gu, Guanghua and Wu, Baoyuan and Zhao, Yao and Wei, Yunchao},
  year = 2025,
  volume = {39},
  pages = {7184--7192},
  copyright = {Copyright (c) 2025 Association for the Advancement of Artificial Intelligence},
  language = {en}
}

@article{arxiv2017kinetics,
  title = {The Kinetics Human Action Video Dataset},
  author = {Kay, Will and Carreira, Joao and Simonyan, Karen and Zhang, Brian and Hillier, Chloe and Vijayanarasimhan, Sudheendra and Viola, Fabio and Green, Tim and Back, Trevor and Natsev, Paul and Suleyman, Mustafa and Zisserman, Andrew},
  year = 2017,
  journal = {arXiv preprint arXiv:1705.06950},
  eprint = {1705.06950},
  pages = {1--22},
  archiveprefix = {arXiv}
}

@misc{arxiv2018vq-vae,
  title = {Neural Discrete Representation Learning},
  author = {van den Oord, Aaron and Vinyals, Oriol and Kavukcuoglu, Koray},
  year = 2018,
  number = {arXiv:1711.00937},
  eprint = {1711.00937},
  primaryclass = {cs},
  archiveprefix = {arXiv},
  note = {arXiv:1711.00937},
}

@misc{arxiv2021clip,
  title = {Learning Transferable Visual Models From Natural Language Supervision},
  author = {Radford, Alec and Kim, Jong Wook and Hallacy, Chris and Ramesh, Aditya and Goh, Gabriel and Agarwal, Sandhini and Sastry, Girish and Askell, Amanda and Mishkin, Pamela and Clark, Jack and Krueger, Gretchen and Sutskever, Ilya},
  year = 2021,
  number = {arXiv:2103.00020},
  eprint = {2103.00020},
  archiveprefix = {arXiv},
  note = {arXiv:2103.00020},
}

@misc{arxiv2022autoencoding,
  title = {Auto-Encoding Variational Bayes},
  author = {Kingma, Diederik P. and Welling, Max},
  year = 2022,
  number = {arXiv:1312.6114},
  eprint = {1312.6114},
  primaryclass = {stat},
  archiveprefix = {arXiv},
  note = {arXiv:1312.6114},
}

@misc{arxiv2022make-a-video,
  title = {Make-A-Video: Text-to-Video Generation without Text-Video Data},
  author = {Singer, Uriel and Polyak, Adam and Hayes, Thomas and Yin, Xi and An, Jie and Zhang, Songyang and Hu, Qiyuan and Yang, Harry and Ashual, Oron and Gafni, Oran and Parikh, Devi and Gupta, Sonal and Taigman, Yaniv},
  year = 2022,
  number = {arXiv:2209.14792},
  eprint = {2209.14792},
  primaryclass = {cs},
  archiveprefix = {arXiv},
  note = {arXiv:2209.14792},
}

@misc{arxiv2023stablevideodiffusion,
  title = {Stable Video Diffusion: Scaling Latent Video Diffusion Models to Large Datasets},
  author = {Blattmann, Andreas and Dockhorn, Tim and Kulal, Sumith and Mendelevitch, Daniel and Kilian, Maciej and Lorenz, Dominik and Levi, Yam and English, Zion and Voleti, Vikram and Letts, Adam and Jampani, Varun and Rombach, Robin},
  year = 2023,
  number = {arXiv:2311.15127},
  eprint = {2311.15127},
  primaryclass = {cs},
  archiveprefix = {arXiv},
  note = {arXiv:2311.15127},
}

@misc{arxiv2023vbench,
  title = {VBench: Comprehensive Benchmark Suite for Video Generative Models},
  author = {Huang, Ziqi and He, Yinan and Yu, Jiashuo and Zhang, Fan and Si, Chenyang and Jiang, Yuming and Zhang, Yuanhan and Wu, Tianxing and Jin, Qingyang and Chanpaisit, Nattapol and Wang, Yaohui and Chen, Xinyuan and Wang, Limin and Lin, Dahua and Qiao, Yu and Liu, Ziwei},
  year = 2023,
  number = {arXiv:2311.17982},
  eprint = {2311.17982},
  primaryclass = {cs},
  archiveprefix = {arXiv},
  note = {arXiv:2311.17982},
}

@misc{arxiv2024demamba,
  title = {DeMamba: AI-Generated Video Detection on Million-Scale GenVideo Benchmark},
  author = {Chen, Haoxing and Hong, Yan and Huang, Zizheng and Xu, Zhuoer and Gu, Zhangxuan and Li, Yaohui and Lan, Jun and Zhu, Huijia and Zhang, Jianfu and Wang, Weiqiang and Li, Huaxiong},
  year = 2024,
  number = {arXiv:2405.19707},
  eprint = {2405.19707},
  primaryclass = {cs},
  archiveprefix = {arXiv},
  note = {arXiv:2405.19707},
}

@misc{arxiv2024genvidbench,
  title = {GenVidBench: A Challenging Benchmark for Detecting AI-Generated Video},
  author = {Ni, Zhen-Liang and Yan, Qiangyu and Yuan, Tianning and Huang, Mouxiao and Hu, Hailin and Chen, Xinghao and Wang, Yunhe},
  year = 2024,
  language = {en}
}

@misc{arxiv2024stablediffusion3,
  title = {Scaling Rectified Flow Transformers for High-Resolution Image Synthesis},
  author = {Esser, Patrick and Kulal, Sumith and Blattmann, Andreas and Entezari, Rahim and M{\"u}ller, Jonas and Saini, Harry and Levi, Yam and Lorenz, Dominik and Sauer, Axel and Boesel, Frederic and Podell, Dustin and Dockhorn, Tim and English, Zion and Lacey, Kyle and Goodwin, Alex and Marek, Yannik and Rombach, Robin},
  year = 2024,
  number = {arXiv:2403.03206},
  eprint = {2403.03206},
  primaryclass = {cs},
  archiveprefix = {arXiv},
  note = {arXiv:2403.03206},
}

@misc{arxiv2025bagel,
  title = {Emerging Properties in Unified Multimodal Pretraining},
  author = {Deng, Chaorui and Zhu, Deyao and Li, Kunchang and Gou, Chenhui and Li, Feng and Wang, Zeyu and Zhong, Shu and Yu, Weihao and Nie, Xiaonan and Song, Ziang and Shi, Guang and Fan, Haoqi},
  year = 2025,
  number = {arXiv:2505.14683},
  eprint = {2505.14683},
  primaryclass = {cs},
  archiveprefix = {arXiv},
  note = {arXiv:2505.14683},
}

@misc{arxiv2025busterx++,
  title = {BusterX++: Towards Unified Cross-Modal AI-Generated Content Detection and Explanation with MLLM},
  author = {Wen, Haiquan and Li, Tianxiao and Huang, Zhenglin and He, Yiwei and Cheng, Guangliang},
  year = 2025,
  number = {arXiv:2507.14632},
  eprint = {2507.14632},
  primaryclass = {cs},
  archiveprefix = {arXiv},
  note = {arXiv:2507.14632},
}

@misc{arxiv2025dinov3,
  title = {DINOv3},
  author = {Sim{\'e}oni, Oriane and Vo, Huy V. and Seitzer, Maximilian and Baldassarre, Federico and Oquab, Maxime and Jose, Cijo and Khalidov, Vasil and Szafraniec, Marc and Yi, Seungeun and Ramamonjisoa, Micha{\"e}l and Massa, Francisco and Haziza, Daniel and Wehrstedt, Luca and Wang, Jianyuan and Darcet, Timoth{\'e}e and Moutakanni, Th{\'e}o and Sentana, Leonel and Roberts, Claire and Vedaldi, Andrea and Tolan, Jamie and Brandt, John and Couprie, Camille and Mairal, Julien and J{\'e}gou, Herv{\'e} and Labatut, Patrick and Bojanowski, Piotr},
  year = 2025,
  number = {arXiv:2508.10104},
  eprint = {2508.10104},
  primaryclass = {cs},
  archiveprefix = {arXiv},
  note = {arXiv:2508.10104},
}

@misc{arxiv2025ivy-fake,
  title = {IVY-FAKE: A Unified Explainable Framework and Benchmark for Image and Video AIGC Detection},
  author = {Zhang, Wayne and Jiang, Changjiang and Zhang, Zhonghao and Si, Chenyang and Yu, Fengchang and Peng, Wei},
  year = 2025,
  number = {arXiv:2506.00979},
  eprint = {2506.00979},
  primaryclass = {cs},
  archiveprefix = {arXiv},
  note = {arXiv:2506.00979},
}

@misc{arxiv2025learning,
  title = {Learning Human-Perceived Fakeness in AI-Generated Videos via Multimodal LLMs},
  author = {Fu, Xingyu and Liu, Siyi and Xu, Yinuo and Lu, Pan and Hu, Guangqiuse and Yang, Tianbo and Anantasagar, Taran and Shen, Christopher and Mao, Yikai and Liu, Yuanzhe and Shah, Keyush and Lee, Chung Un and Choi, Yejin and Zou, James and Roth, Dan and {Callison-Burch}, Chris},
  year = 2025,
  number = {arXiv:2509.22646},
  eprint = {2509.22646},
  primaryclass = {cs},
  archiveprefix = {arXiv},
  note = {arXiv:2509.22646},
}

@misc{arxiv2025loki,
  title = {LOKI: A Comprehensive Synthetic Data Detection Benchmark Using Large Multimodal Models},
  author = {Ye, Junyan and Zhou, Baichuan and Huang, Zilong and Zhang, Junan and Bai, Tianyi and Kang, Hengrui and He, Jun and Lin, Honglin and Wang, Zihao and Wu, Tong and Wu, Zhizheng and Chen, Yiping and Lin, Dahua and He, Conghui and Li, Weijia},
  year = 2025,
  number = {arXiv:2410.09732},
  eprint = {2410.09732},
  primaryclass = {cs},
  archiveprefix = {arXiv},
  note = {arXiv:2410.09732},
}

@misc{arxiv2025saga,
  title = {SAGA: Source Attribution of Generative AI Videos},
  author = {Kundu, Rohit and Mohanty, Vishal and Xiong, Hao and Jia, Shan and Balachandran, Athula and {Roy-Chowdhury}, Amit K.},
  year = 2025,
  number = {arXiv:2511.12834},
  eprint = {2511.12834},
  primaryclass = {cs},
  archiveprefix = {arXiv},
  note = {arXiv:2511.12834},
}

@misc{arxiv2025seedance1.0,
  title = {Seedance 1.0: Exploring the Boundaries of Video Generation Models},
  author = {Gao, Yu and Guo, Haoyuan and Hoang, Tuyen and Huang, Weilin and Jiang, Lu and Kong, Fangyuan and Li, Huixia and Li, Jiashi and Li, Liang and Li, Xiaojie and Li, Xunsong and Li, Yifu and Lin, Shanchuan and Lin, Zhijie and Liu, Jiawei and Liu, Shu and Nie, Xiaonan and Qing, Zhiwu and Ren, Yuxi and Sun, Li and Tian, Zhi and Wang, Rui and Wang, Sen and Wei, Guoqiang and Wu, Guohong and Wu, Jie and Xia, Ruiqi and Xiao, Fei and Xiao, Xuefeng and Yan, Jiangqiao and Yang, Ceyuan and Yang, Jianchao and Yang, Runkai and Yang, Tao and Yang, Yihang and Ye, Zilyu and Zeng, Xuejiao and Zeng, Yan and Zhang, Heng and Zhao, Yang and Zheng, Xiaozheng and Zhu, Peihao and Zou, Jiaxin and Zuo, Feilong},
  year = 2025,
  copyright = {arXiv.org perpetual, non-exclusive license},
  language = {en}
}

@misc{arxiv2025skyra,
  title = {Skyra: AI-Generated Video Detection via Grounded Artifact Reasoning},
  author = {Li, Yifei and Zheng, Wenzhao and Zhang, Yanran and Sun, Runze and Zheng, Yu and Chen, Lei and Zhou, Jie and Lu, Jiwen},
  year = 2025,
  number = {arXiv:2512.15693},
  eprint = {2512.15693},
  primaryclass = {cs},
  archiveprefix = {arXiv},
  note = {arXiv:2512.15693},
}

@misc{arxiv2025vidguard-r1,
  title = {VidGuard-R1: AI-Generated Video Detection and Explanation via Reasoning MLLMs and RL},
  author = {Park, Kyoungjun and Yang, Yifan and Yi, Juheon and Zheng, Shicheng and Shen, Yifei and Han, Dongqi and Shan, Caihua and Muaz, Muhammad and Qiu, Lili},
  year = 2025,
  number = {arXiv:2510.02282},
  eprint = {2510.02282},
  primaryclass = {cs},
  archiveprefix = {arXiv},
  note = {arXiv:2510.02282},
}

@misc{arxiv2025wan,
  title = {Wan: Open and Advanced Large-Scale Video Generative Models},
  author = {WanTeam and Wang, Ang and Ai, Baole and Wen, Bin and Mao, Chaojie and Xie, Chen-Wei and Chen, Di and Yu, Feiwu and Zhao, Haiming and Yang, Jianxiao and Zeng, Jianyuan and Wang, Jiayu and Zhang, Jingfeng and Zhou, Jingren and Wang, Jinkai and Chen, Jixuan and Zhu, Kai and Zhao, Kang and Yan, Keyu and Huang, Lianghua and Feng, Mengyang and Zhang, Ningyi and Li, Pandeng and Wu, Pingyu and Chu, Ruihang and Feng, Ruili and Zhang, Shiwei and Sun, Siyang and Fang, Tao and Wang, Tianxing and Gui, Tianyi and Weng, Tingyu and Shen, Tong and Lin, Wei and Wang, Wei and Wang, Wei and Zhou, Wenmeng and Wang, Wente and Shen, Wenting and Yu, Wenyuan and Shi, Xianzhong and Huang, Xiaoming and Xu, Xin and Kou, Yan and Lv, Yangyu and Li, Yifei and Liu, Yijing and Wang, Yiming and Zhang, Yingya and Huang, Yitong and Li, Yong and Wu, You and Liu, Yu and Pan, Yulin and Zheng, Yun and Hong, Yuntao and Shi, Yupeng and Feng, Yutong and Jiang, Zeyinzi and Han, Zhen and Wu, Zhi-Fan and Liu, Ziyu},
  year = 2025,
  number = {arXiv:2503.20314},
  eprint = {2503.20314},
  primaryclass = {cs},
  archiveprefix = {arXiv},
  note = {arXiv:2503.20314},
}

@inproceedings{cvpr2020cnngenerated,
  title = {CNN-Generated Images Are Surprisingly Easy to Spot... for Now},
  booktitle = CVPR,
  author = {Wang, Sheng-Yu and Wang, Oliver and Zhang, Richard and Owens, Andrew and Efros, Alexei A.},
  year = 2020,
  pages = {8695--8704}
}

@inproceedings{cvpr2020watchyourup-convolution,
  title = {Watch Your Up-Convolution: CNN Based Generative Deep Neural Networks Are Failing to Reproduce Spectral Distributions},
  booktitle = CVPR,
  author = {Durall, Ricard and Keuper, Margret and Keuper, Janis},
  year = 2020,
  eprint = {2003.01826},
  primaryclass = {cs},
  archiveprefix = {arXiv}
}

@inproceedings{cvpr2022stablediffusion,
  title = {High-Resolution Image Synthesis With Latent Diffusion Models},
  booktitle = CVPR,
  author = {Rombach, Robin and Blattmann, Andreas and Lorenz, Dominik and Esser, Patrick and Ommer, Bj{\"o}rn},
  year = 2022,
  pages = {10684--10695},
  language = {en}
}

@inproceedings{cvpr2023unifd,
  title = {Towards Universal Fake Image Detectors That Generalize Across Generative Models},
  booktitle = CVPR,
  author = {Ojha, Utkarsh and Li, Yuheng and Lee, Yong Jae},
  year = 2023,
  pages = {24480--24489},
  language = {en}
}

@inproceedings{cvpr2024fatformer,
  title = {Forgery-Aware Adaptive Transformer for Generalizable Synthetic Image Detection},
  booktitle = CVPR,
  author = {Liu, Huan and Tan, Zichang and Tan, Chuangchuang and Wei, Yunchao and Wang, Jingdong and Zhao, Yao},
  year = 2024,
  pages = {10770--10780},
  language = {en}
}

@inproceedings{cvpr2024npr,
  title = {Rethinking the Up-Sampling Operations in CNN-Based Generative Network for Generalizable Deepfake Detection},
  booktitle = CVPR,
  author = {Tan, Chuangchuang and Zhao, Yao and Wei, Shikui and Gu, Guanghua and Liu, Ping and Wei, Yunchao},
  year = 2024,
  pages = {28130--28139},
  language = {en}
}

@inproceedings{cvpr2025b-free,
  title = {A Bias-Free Training Paradigm for More General AI-Generated Image Detection},
  booktitle = CVPR,
  author = {Guillaro, Fabrizio and Zingarini, Giada and Usman, Ben and Sud, Avneesh and Cozzolino, Davide and Verdoliva, Luisa},
  year = 2025,
  eprint = {2412.17671},
  primaryclass = {cs},
  archiveprefix = {arXiv}
}

@inproceedings{cvpr2025co-spy,
  title = {CO-SPY: Combining Semantic and Pixel Features to Detect Synthetic Images by AI},
  booktitle = CVPR,
  author = {Cheng, Siyuan and Lyu, Lingjuan and Wang, Zhenting and Zhang, Xiangyu and Sehwag, Vikash},
  year = 2025,
  eprint = {2503.18286},
  primaryclass = {cs},
  archiveprefix = {arXiv}
}

@inproceedings{cvpr2025secretliesincolor,
  title = {Secret Lies in Color: Enhancing AI-Generated Images Detection with Color Distribution Analysis},
  booktitle = CVPR,
  author = {Jia, Zexi and Huang, Chuanwei and Zhu, Yeshuang and Fei, Hongyan and Duan, Xiaoyue and Yuan, Zhiqiang and Deng, Ying and Zhang, Jiapei and Zhang, Jinchao and Zhou, Jie},
  year = 2025,
  pages = {13445--13454},
  language = {en}
}

@inproceedings{cvpr2025unite,
  title = {Towards a Universal Synthetic Video Detector: From Face or Background Manipulations to Fully AI-Generated Content},
  booktitle = CVPR,
  author = {Kundu, Rohit and Xiong, Hao and Mohanty, Vishal and Balachandran, Athula and {Roy-Chowdhury}, Amit K.},
  year = 2025,
  eprint = {2412.12278},
  primaryclass = {cs},
  archiveprefix = {arXiv}
}

@inproceedings{cvprw2024raising,
  title = {Raising the Bar of AI-Generated Image Detection with CLIP},
  booktitle = CVPRW,
  author = {Cozzolino, Davide and Poggi, Giovanni and Corvi, Riccardo and Nie{\ss}ner, Matthias and Verdoliva, Luisa},
  year = 2024,
  eprint = {2312.00195},
  primaryclass = {cs},
  archiveprefix = {arXiv}
}

@inproceedings{eccv2024rine,
  title = {Leveraging Representations from Intermediate Encoder-Blocks for Synthetic Image Detection},
  booktitle = ECCV,
  author = {Koutlis, Christos and Papadopoulos, Symeon},
  year = 2024
}

@inproceedings{eccv2024zeroshot,
  title = {Zero-Shot Detection of AI-Generated Images},
  booktitle = ECCV,
  author = {Cozzolino, Davide and Poggi, Giovanni and Nie{\ss}ner, Matthias and Verdoliva, Luisa},
  year = 2024,
  eprint = {2409.15875},
  primaryclass = {cs},
  archiveprefix = {arXiv}
}

@inproceedings{iccv2023dire,
  title = {DIRE for Diffusion-Generated Image Detection},
  booktitle = ICCV,
  author = {Wang, Zhendong and Bao, Jianmin and Zhou, Wengang and Wang, Weilun and Hu, Hezhen and Chen, Hong and Li, Houqiang},
  year = 2023,
  pages = {22445--22455},
  language = {en}
}

@inproceedings{iccv2023dit,
  title = {Scalable Diffusion Models with Transformers},
  booktitle = ICCV,
  author = {Peebles, William and Xie, Saining},
  year = 2023,
  eprint = {2212.09748},
  primaryclass = {cs},
  archiveprefix = {arXiv}
}

@inproceedings{iccv2025d^3qe,
  title = {\$\textbackslash bf\textbraceleft D\textasciicircum 3\textbraceright\$QE: Learning Discrete Distribution Discrepancy-Aware Quantization Error for Autoregressive-Generated Image Detection},
  booktitle = ICCV,
  author = {Zhang, Yanran and Yu, Bingyao and Zheng, Yu and Zheng, Wenzhao and Duan, Yueqi and Chen, Lei and Zhou, Jie and Lu, Jiwen},
  year = 2025,
  eprint = {2510.05891},
  primaryclass = {cs},
  archiveprefix = {arXiv}
}

@inproceedings{iccv2025d3,
  title = {D3: Training-Free AI-Generated Video Detection Using Second-Order Features},
  booktitle = ICCV,
  author = {Zheng, Chende and {suo}, Ruiqi and Lin, Chenhao and Zhao, Zhengyu and Yang, Le and Liu, Shuai and Yang, Minghui and Wang, Cong and Shen, Chao},
  year = 2025,
  eprint = {2508.00701},
  primaryclass = {cs},
  archiveprefix = {arXiv}
}

@inproceedings{iclr2018progan,
  title = {Progressive Growing of GANs for Improved Quality, Stability, and Variation},
  booktitle = ICLR,
  author = {Karras, Tero and Aila, Timo and Laine, Samuli and Lehtinen, Jaakko},
  year = 2018,
  eprint = {1710.10196},
  primaryclass = {cs},
  archiveprefix = {arXiv}
}

@inproceedings{iclr2025aligned,
  title = {Aligned Datasets Improve Detection of Latent Diffusion-Generated Images},
  booktitle = ICLR,
  author = {Rajan, Anirudh Sundara and Ojha, Utkarsh and Schloesser, Jedidiah and Lee, Yong Jae},
  year = 2025,
  language = {en}
}

@inproceedings{iclr2025sanity,
  title = {A Sanity Check for AI-Generated Image Detection},
  booktitle = ICLR,
  author = {Yan, Shilin and Li, Ouxiang and Cai, Jiayin and Hao, Yanbin and Jiang, Xiaolong and Hu, Yao and Xie, Weidi},
  year = 2025,
  language = {en}
}

@inproceedings{icml2024drct,
  title = {DRCT: Diffusion Reconstruction Contrastive Training towards Universal Detection of Diffusion Generated Images},
  booktitle = ICML,
  author = {Chen, Baoying and Zeng, Jishen and Yang, Jianquan and Yang, Rui},
  year = 2024,
  language = {en}
}

@inproceedings{icml2025effort,
  ids = {arxiv2024effort},
  title = {Orthogonal Subspace Decomposition for Generalizable AI-Generated Image Detection},
  booktitle = ICML,
  author = {Yan, Zhiyuan and Wang, Jiangming and Jin, Peng and Zhang, Ke-Yue and {Chengchun Liu} and Chen, Shen and Yao, Taiping and Ding, Shouhong and Wu, Baoyuan and Yuan, Li},
  year = 2025,
  eprint = {2411.15633},
  primaryclass = {cs},
  archiveprefix = {arXiv}
}

@inproceedings{kdd2025safe,
  ids = {arxiv2024improvingsyntheticimagedetectiontowardsgeneralization},
  title = {Improving Synthetic Image Detection Towards Generalization: An Image Transformation Perspective},
  booktitle = {KDD},
  author = {Li, Ouxiang and Cai, Jiayin and Hao, Yanbin and Jiang, Xiaolong and Hu, Yao and Feng, Fuli},
  year = 2025,
  eprint = {2408.06741},
  primaryclass = {cs},
  archiveprefix = {arXiv}
}

@inproceedings{nips2014gan,
  title = {Generative Adversarial Networks},
  booktitle = NIPS,
  author = {Goodfellow, Ian J. and {Pouget-Abadie}, Jean and Mirza, Mehdi and Xu, Bing and {Warde-Farley}, David and Ozair, Sherjil and Courville, Aaron and Bengio, Yoshua},
  year = 2014,
  eprint = {1406.2661},
  primaryclass = {stat},
  archiveprefix = {arXiv}
}

@inproceedings{nips2020ddpm,
  title = {Denoising Diffusion Probabilistic Models},
  booktitle = NIPS,
  author = {Ho, Jonathan and Jain, Ajay and Abbeel, Pieter},
  year = 2020,
  eprint = {2006.11239},
  primaryclass = {cs},
  archiveprefix = {arXiv}
}

@inproceedings{nips2023genimage,
  title = {GenImage: A Million-Scale Benchmark for Detecting AI-Generated Image},
  booktitle = NIPS,
  author = {Zhu, Mingjian and Chen, Hanting and Yan, Qiangyu and Huang, Xudong and Lin, Guanyu and Li, Wei and Tu, Zhijun and Hu, Hailin and Hu, Jie and Wang, Yunhe},
  year = 2023,
  language = {en}
}

@inproceedings{nips2024var,
  title = {Visual Autoregressive Modeling: Scalable Image Generation via Next-Scale Prediction},
  booktitle = NIPS,
  author = {Tian, Keyu and Jiang, Yi and Yuan, Zehuan and Peng, Bingyue and Wang, Liwei},
  year = 2024,
  eprint = {2404.02905},
  primaryclass = {cs},
  archiveprefix = {arXiv}
}

@inproceedings{nips2025dda,
  title = {Dual Data Alignment Makes AI-Generated Image Detector Easier Generalizable},
  booktitle = NIPS,
  author = {Chen, Ruoxin and Xi, Junwei and Yan, Zhiyuan and Zhang, Ke-Yue and Wu, Shuang and Xie, Jingyi and Chen, Xu and Xu, Lei and Guan, Isabel and Yao, Taiping and Ding, Shouhong},
  year = 2025,
  language = {en}
}

@inproceedings{nips2025nsg-vd,
  title = {Physics-Driven Spatiotemporal Modeling for AI-Generated Video Detection},
  booktitle = NIPS,
  author = {Zhang, Shuhai and Lian, ZiHao and Yang, Jiahao and Li, Daiyuan and Pang, Guoxuan and Liu, Feng and Han, Bo and Li, Shutao and Tan, Mingkui},
  year = 2025,
  language = {en}
}

@inproceedings{nips2025seeingwhatmatters,
  title = {Seeing What Matters: Generalizable AI-Generated Video Detection with Forensic-Oriented Augmentation},
  booktitle = NIPS,
  author = {Corvi, Riccardo and Cozzolino, Davide and Prashnani, Ekta and Mello, Shalini De and Nagano, Koki and Verdoliva, Luisa},
  year = 2025,
  eprint = {2506.16802},
  primaryclass = {cs},
  archiveprefix = {arXiv}
}

@article{oquab2023dinov2,
  title = {DINOv2: Learning Robust Visual Features without Supervision},
  author = {Oquab, Maxime and Darcet, Timoth{\'e}e and Moutakanni, Th{\'e}o and Vo, Huy V. and Szafraniec, Marc and Khalidov, Vasil and Fernandez, Pierre and Haziza, Daniel and Massa, Francisco and {El-Nouby}, Alaaeldin and Assran, Mido and Ballas, Nicolas and Galuba, Wojciech and Howes, Russell and Huang, Po-Yao and Li, Shang-Wen and Misra, Ishan and Rabbat, Michael and Sharma, Vasu and Synnaeve, Gabriel and Xu, Hu and Jegou, Herve and Mairal, Julien and Labatut, Patrick and Joulin, Armand and Bojanowski, Piotr},
  year = 2023,
  journal = {Transactions on Machine Learning Research},
  issn = {2835-8856},
  language = {en}
}

@misc{wang2023allseeing,
  title = {The All-Seeing Project: Towards Panoptic Visual Recognition and Understanding of the Open World},
  author = {Wang, Weiyun and Shi, Min and Li, Qingyun and Wang, Wenhai and Huang, Zhenhang and Xing, Linjie and Chen, Zhe and Li, Hao and Zhu, Xizhou and Cao, Zhiguo and Chen, Yushi and Lu, Tong and Dai, Jifeng and Qiao, Yu},
  year = 2023,
  number = {arXiv:2308.01907},
  eprint = {2308.01907},
  primaryclass = {cs},
  archiveprefix = {arXiv},
  note = {arXiv:2308.01907},
}
\bibliographystyle{unsrtnat}

%%%%%%%%%%%%%%%%%%%%%%%%%%%%%%%%%%%%%%%%%%%%%%%%%%%%%%%%%%%%
% \appendix
\newpage
\appendix
\onecolumn

% \begin{left}
{\LARGE \bfseries Appendix}
% \end{center}

The appendix is organized as follows.
\begin{itemize}
    \item \textbf{\Cref{sec:appendix_dataset_analysis}: Dataset Analysis.} Additional frequency-domain analysis of image and video datasets.
    \item \textbf{\Cref{sec:appendix_implementation}: Implementation Details.} Baseline checkpoints, preprocessing, and inference settings.
    \item \textbf{\Cref{sec:appendix_benchmark}: Detailed Benchmark Results.} Extended results on image- and video-based AIGC detection benchmarks.
    \item \textbf{\Cref{sec:appendix_ablation}: More Ablation Studies.} Additional analyses of data sources, tuning, image transforms, and CM-SupCon loss weights.
    \item \textbf{\Cref{sec:appendix_scaling}: Scaling Potentials.} Temporal scaling results.
    \item \textbf{\Cref{sec:appendix_discussion}: Discussion.} Broader impacts and deployment risks.
\end{itemize}

\begin{figure}[h]
    \centering
    \includegraphics[width=0.75\linewidth]{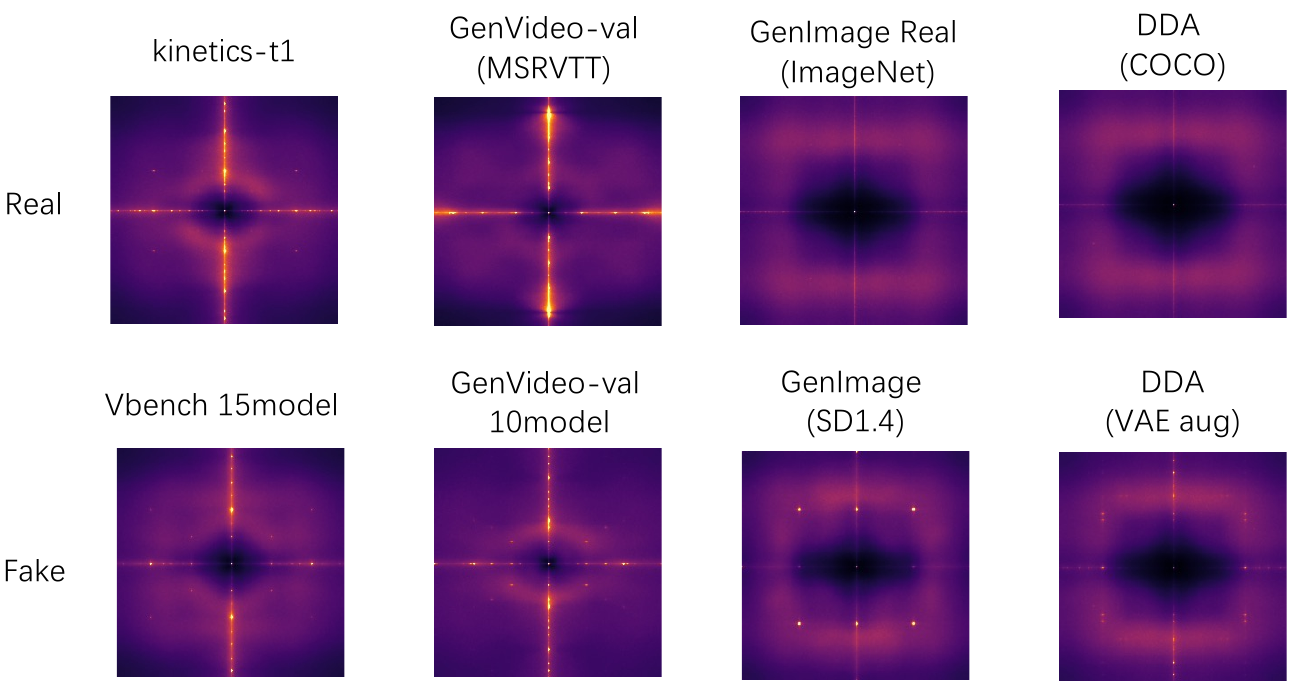}
    \caption{\textbf{2D power spectra of reconstructed noise.} Spectra are averaged over 1,000 samples. Videos exhibit a distinct axial cross pattern, in contrast to the broader spectral spread of images. Generated images show periodic grid artifacts, while generated videos display a different configuration of spectral anomalies. This highlights a fundamental divergence in the frequency-space fingerprints of video and image generative models.}
    \label{fig:analysis_dft}
\end{figure}

\begin{figure}[h]
    \centering
    \includegraphics[width=0.75\linewidth]{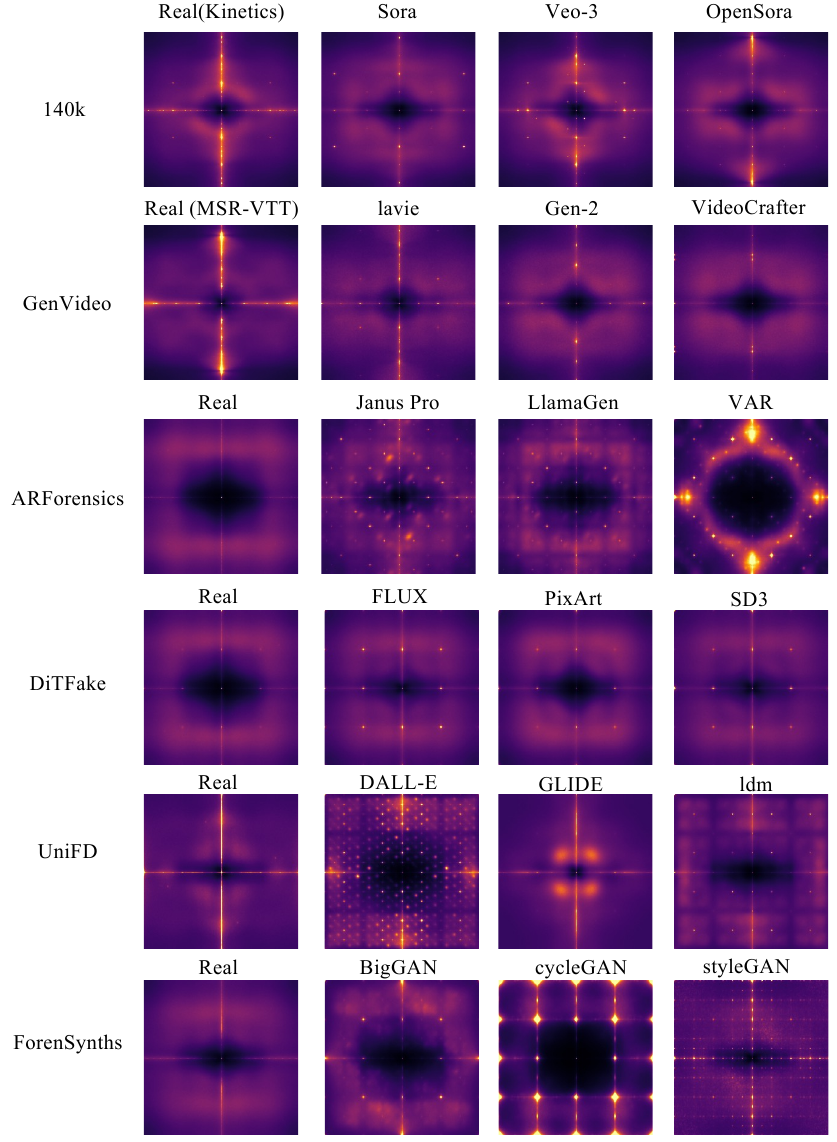}
    \caption{\textbf{Average Fourier power spectra across different generative models and datasets.} The visualizations compare the frequency-domain characteristics of real data (leftmost column) against various image and video generation frameworks. Note the unique structural artifacts (e.g., grid patterns) present in the generated samples across different model families.}
    \label{fig:spectrum_model}
\end{figure}

\section{Dataset Analysis}
\label{sec:appendix_dataset_analysis}

This section includes the spectral visualizations in \cref{fig:analysis_dft,fig:spectrum_model}.

The spectral analysis in \cref{fig:analysis_dft,fig:spectrum_model} reveals a fundamental divergence between the noise distributions of image and video modalities. We follow the implementation in~\cite{nips2025seeingwhatmatters} and randomly select 1k samples from each subset. In real-world data, video exhibits distinct frequency characteristics, shaped by temporal factors like motion blur and artifacts from video compression codecs (e.g., H.264/H.265). Beyond natural content, we observe that generative models imprint distinct architectural traces in the Fourier domain. Critically, a significant gap exists between the spectral patterns of video generation models (e.g., Sora, Gen-2) and image-based Diffusion Transformers (e.g., FLUX, SD3). While image generators often produce periodic, grid-like artifacts, video models induce more complex spatio-temporal frequency distortions. This discrepancy implies that forensic models trained solely on image-based synthetic data face an inherent generalization bottleneck when applied to video, due to this pronounced distribution shift. Consequently, we posit that a joint training paradigm is essential for building a robust and universal detection framework.

\section{Implementation Details}
\label{sec:appendix_implementation}

This section documents the evaluation settings used for the main and appendix comparisons in \cref{tab:hybrid_acc,tab:in_the_wild,tab:hybrid_ap}.

\begin{enumerate}
    \item \textbf{WaveRep}~\cite{nips2025seeingwhatmatters} trains DINO classifiers toward intrinsic low-level artifacts through wavelet-based data augmentation and discriminative feature analysis to enhance generalizability for AIGV detection. We use their released weights and adopt the same 504p crop preprocessing method as their original implementation.
    \item \textbf{Effort}~\cite{icml2025effort} is a parameter-efficient method that leverages SVD to construct orthogonal subspaces. This design prevents catastrophic forgetting of pre-trained semantic knowledge while efficiently learning forgery patterns. Our implementation uses their released weights trained on SD1.4 with identical preprocessing.
    \item \textbf{NPR}~\cite{cvpr2024npr} distinguishes synthetic images by analyzing low-level neighboring pixel relationships, which are characteristic of upsampling patterns in AI-generated content. It implements this approach by training a ResNet-50 model for identification. We use their official checkpoint for evaluation.
    \item \textbf{RINE}~\cite{eccv2024rine} improves AI-generated image detection by mapping intermediate CLIP features to a forgery-aware vector space. We utilize the author-released weights trained on the ProGAN 4class dataset and adopt the same resize preprocessing method as the original implementation.
    \item \textbf{B-Free}~\cite{cvpr2025b-free} employs a self-conditioned diffusion reconstruction paradigm to enforce semantic alignment between real and synthetic images, thereby isolating differences to generation artifacts. The method incorporates content augmentation via inpainting and fine-tunes a DINOv2+reg ViT using large crops to preserve forensic signals. We utilize their provided checkpoint, applying the same 504p five-crop preprocessing.
    \item \textbf{CO-SPY}~\cite{cvpr2025co-spy} enhances semantic and artifact features for robust synthetic image detection. Following the official code, we evaluate the two author-released implementations trained on SD1.4 and ProGAN.
    \item \textbf{DDA}~\cite{nips2025dda} aligns synthetic and real images in both pixel and frequency domains to mitigate spurious correlations and improve detector generalization. For evaluation, we utilize the author-released weights and identical settings, with the sole exception that we forgo applying the same JPEG Q=96 compression on the test sets to ensure a fair comparison with other methods. Our results show that applying or forgoing this specific JPEG compression has a negligible impact on the performance of both the DDA method and our approach.
\end{enumerate}

\section{Detailed Benchmark Results}
\label{sec:appendix_benchmark}

This section includes the benchmark overview and detailed results in \cref{tab:dataset-overview,tab:hybrid_ap,tab:genimage,tab:forensyths,tab:unifd,tab:ditfake,tab:arforensics,tab:magic-acc,tab:genvideo_full}.

\begin{table}[t]
    \caption{\textbf{Overview of the evaluation benchmarks.} Our evaluation covers 14 benchmarks (4 video benchmarks, 5 image benchmarks, and 5 in-the-wild benchmarks), spanning a diverse range of generator architectures to ensure broad and comprehensive assessment.}
  \label{tab:dataset-overview}
  \centering
  % \resizebox{\linewidth}{!}{
    \begin{tabular}{cccc}
      \toprule
      Dataset & Type & Real/Fake & Model Types \\
      \midrule
      Magic Videos~\cite{iclr2026preserving} & Video & 0.5k/1.5k & DiT, Commercial \\
      GenVideo~\cite{arxiv2024demamba} & Video & 10k/9.5k & SVD, DiT \\
      Genbuster++~\cite{arxiv2025busterx++} & Video & 1k/1k & Commercial \\
      DeepTraceReward~\cite{arxiv2025learning} & Video & 3.3k/3.3k & Commercial \\
      \midrule
      ForenSynths~\cite{cvpr2020cnngenerated} & Image & 45k/45k & GAN,Deepfake \\
      UniFD~\cite{cvpr2023unifd} & Image & 8k/8k & LDM \\
      DiTFake~\cite{kdd2025safe} & Image & 15k/15k & DiT \\
      ARforensics~\cite{iccv2025d^3qe} & Image & 39k/39k & Auto-Regressive \\
      GenImage~\cite{nips2023genimage} & Image & 48K/48K & SD,GAN \\
      \midrule
      Chameleon~\cite{iclr2025sanity} & in-the-wild & 14.9k/11.2k & in-the-wild \\
      SynthWildx~\cite{cvprw2024raising} & in-the-wild & 500/1.5K & X  \\
      WildRF~\cite{arxiv2024realtime} & in-the-wild & 500/500 & unknown \\
      AIGIBench~\cite{nips2025aigibench} & in-the-wild & 9k/9k & SocialRF, CommunityAI\\
      RR-Dataset~\cite{iccv2025rrdataset} & in-the-wild & 25k/25k & SD3.5, FLUX.1 \\
      \bottomrule
    \end{tabular}
  % }
\end{table}

\begin{table}[htbp]
    \centering
    \caption{\textbf{Comparison results on 4 video-based and 4 image-based AIGC detection benchmarks in AP(\%)}. Except for DeMamba, for which no public checkpoint is available and which we retrain on GenVideo-100k, all results are obtained by us using the authors’ released pretrained weights and the corresponding inference-time transform settings.}
    \label{tab:hybrid_ap}
    \resizebox{\linewidth}{!}{
    \begin{tabular}{@{}ll
    >{\columncolor[HTML]{E5F6FF}}r cccc
    >{\columncolor[HTML]{FFF9E3}}r cccc
    >{\columncolor[HTML]{FDEBFF}}r}
    \toprule
     &  & \multicolumn{5}{c}{\cellcolor[HTML]{E5F6FF}Video} & \multicolumn{5}{c}{\cellcolor[HTML]{FFF9E3}Image} & \multicolumn{1}{c}{\cellcolor[HTML]{FDEBFF}ALL}     \\ 
    Method & \multicolumn{1}{c}{Type} & \multicolumn{1}{l}{\cellcolor[HTML]{E5F6FF}AVG} & \multicolumn{1}{l}{Magic} & \multicolumn{1}{l}{GenVideo} & \multicolumn{1}{l}{\makecell{GenBu\\ster++}} & \multicolumn{1}{l}{\makecell{DeepTrace\\Reward}} & \multicolumn{1}{l}{\cellcolor[HTML]{FFF9E3}AVG} & \multicolumn{1}{l}{\makecell{ForenSynths\\(GAN)}} & \multicolumn{1}{l}{\begin{tabular}[c]{@{}l@{}}UniFD\\(LDM)\end{tabular}} & \multicolumn{1}{l}{DiTFake} & \multicolumn{1}{l}{\makecell{ARForen\\sics}} & \multicolumn{1}{l}{\cellcolor[HTML]{FDEBFF}\begin{tabular}[c]{@{}l@{}}AVG\end{tabular}} \\
    \midrule
    DeMamba~\cite{arxiv2024demamba} (100k) & Video & 84.21 & 73.78 & 96.66 & 68.49 & 97.90 & 65.40 & 56.50 & 57.00 & 84.96 & 63.14 & 74.81 \\
    Qwen2.5-ViT~\cite{iclr2026preserving} & Video & \underline{95.80} & 96.94 & 99.52 & 87.86 & 98.89 & 71.17 & 52.64 & 62.38 & 95.16 & 74.48 & 83.48 \\
    WaveRep~\cite{nips2025seeingwhatmatters} & Video & 90.09 & 88.99 & 98.18 & 73.92 & 99.26 & 96.01 & 98.87 & 93.83 & 98.91 & 92.42 & 93.05  \\
    \midrule
    NPR~\cite{cvpr2024npr} & Image & 67.83 & 70.19 & 65.26 & 68.89 & 66.96 & 84.34 & 78.85 & 96.22 & 83.37 & 78.93 & 76.08 \\
    FatFormer~\cite{cvpr2024fatformer} & Image & 53.98 & 40.19 & 63.48 & 52.03 & 60.23 & 88.46 & 97.88 & 98.56 & 70.03 & 87.37 & 71.22 \\
    RINE~\cite{eccv2024rine} & Image & 61.73 & 35.48 & 91.06 & 71.99 & 48.38 & 96.77 & 98.80 & 98.96 & 92.45 & 96.85 & 79.25 \\
    Effort~\cite{icml2025effort} & Image & 80.67 & 74.05 & 96.86 & 67.78 & 83.99 & 89.07 & 78.30 & 96.79 & 91.79 & 89.41 & 84.87 \\
    Co-Spy~\cite{cvpr2025co-spy} & Image & 83.41 & 71.16 & 87.16 & 82.77 & 92.56 & 86.01 & 70.69 & 84.21 & 99.30 & 89.83 & 84.71 \\
    Co-Spy (ProGAN) & Image & 62.31 & 51.41 & 58.08 & 75.43 & 64.30 & 93.97 & 87.79 & 98.68 & 96.68 & 92.72 & 78.14  \\
    DDA~\cite{nips2025dda} & Image & 87.87 & 85.47 & 92.75 & 80.57 & 92.69 & 94.79 & 89.06 & 93.59 & 99.89 & 96.61 & 91.33  \\
    B-Free~\cite{cvpr2025b-free} & Image & 86.45 & 80.37 & 98.06 & 77.50 & 89.88 & 97.00 & 94.04 & 96.23 & 99.43 & 98.28 & 91.72  \\
    \midrule
    VINA (Pika+ProGAN) & V+I & 90.78 & 79.25 & 96.64 & 88.91 & 98.33 & \underline{98.55} & 97.69 & 98.74 & 99.54 & 98.22 & \underline{94.67} \\
    VINA (GenVideo+SD1.4) & V+I & 90.58 & 79.60 & 98.88 & 85.21 & 98.63 & 98.49 & 97.35 & 98.36 & 99.93 & 98.32 & 94.54 \\
    VINA (140k+DDA) & V+I & \textbf{97.70} & 95.10 & 99.26 & 96.53 & 99.89 & \textbf{99.08} & 97.77 & 99.01 & 99.95 & 99.60  & \textbf{98.39}  \\
    \bottomrule
    \end{tabular}
    }
\end{table}

\paragraph{Performance Comparison in AP.} \Cref{tab:hybrid_ap} complements the ACC comparison in \cref{tab:hybrid_acc} by evaluating ranking quality across the same video and image benchmarks. Video-specific detectors such as DeMamba and Qwen2.5-ViT achieve higher AP on video benchmarks than on image datasets, while image-based detectors show the opposite tendency and often lose discriminative ranking on videos. In contrast, joint video-image training produces consistently high AP across both modalities. The strongest VINA model achieves 97.70\% video AP, 99.08\% image AP, and 98.39\% overall AP, indicating that the proposed framework improves not only threshold-dependent accuracy but also cross-domain ranking robustness.

\paragraph{Performance on GenImage.} \Cref{tab:genimage} presents a comprehensive benchmark comparison on the GenImage dataset~\cite{nips2023genimage}, evaluating VINA against several state-of-the-art detectors across eight diverse generative models. We evaluate all methods using their released checkpoints and the corresponding test transformations. VINA achieves strong performance without JPEG alignment (94.74\% AVG) and further improves to 95.53\% AVG with the JPEG setting, outperforming the strongest image-only baseline (DDA, 88.98\%) by 6.55\%. Notably, the JPEG variant improves robustness on GLIDE, VQDM, and BigGAN while maintaining near-perfect detection accuracy on the Stable Diffusion and Wukong subsets. The fact that JPEG compression further improves VINA, rather than degrading it, suggests that joint video-image training suppresses shortcut learning from format artifacts and learns more robust generative cues.

\begin{table*}[ht]
    \caption{\textbf{Benchmarking results on GenImage~\cite{nips2023genimage} in terms of ACC(\%).}}
    \label{tab:genimage}
    \centering
    \resizebox{\linewidth}{!}{
    \begin{tabular}{lc|ccccccccc}
    \toprule
    Method & Type & Midjourney & SDv1.4 & SDv1.5 & ADM & GLIDE & Wukong & VQDM & BigGAN & AVG \\
    \midrule
    WaveRep~\cite{nips2025seeingwhatmatters} & V & 80.57 & 82.75 & 82.76 & 75.47 & 78.33 & 82.17 & 78.20 & 82.63 & 80.36 \\
    NPR~\cite{cvpr2024npr} & I & 77.73 & 78.23 & 78.83 & 74.22 & 78.87 & 72.40 & 77.84 & 81.28 & 77.43 \\
    FatFormer~\cite{cvpr2024fatformer} & I & 56.08 & 67.83 & 68.06 & 78.44 & 88.03 & 73.06 & 86.88 & 96.79 & 76.90 \\
    RINE~\cite{eccv2024rine} & I & 57.12 & 83.94 & 83.36 & 74.58 & 80.71 & 84.94 & 89.76 & 94.83 & 81.16  \\
    Effort~\cite{icml2025effort} & I & 76.17 & 87.77 & 87.61 & 70.88 & 80.82 & 86.81 & 82.62 & 75.43 & 81.01 \\
    % B-Free~\cite{cvpr2025b-free} & I & 63.46 & 82.18 & 81.86 & 72.15 & 81.32 & 86.68 & 86.88 & 74.35 & 78.61 \\
    CO-SPY~\cite{cvpr2025co-spy} & I & 69.87 & 93.03 & 92.9 & 58.15 & 81.06 & 91.03 & 71.96 & 64.22 & 77.78 \\
    CO-SPY (ProGAN) & I & 86.98 & 85.40 & 85.58 & 78.79 & 91.26 & 80.28 & 84.86 & 93.57 & 85.84 \\
    DDA~\cite{nips2025dda} & I & 96.00 & 98.57 & 98.53 & 88.03 & 86.13 & 98.61 & 71.83 & 74.11 & 88.98 \\
    \midrule
    VINA (140k+DDA) & V+I & 96.11 & 99.89 & 99.85 & 76.23 & 94.65 & 99.83 & 94.98 & 96.38 & \underline{94.74} \\
    VINA (140k+DDA, JPEG) & V+I & 93.83 & 99.86 & 99.83 & 76.79 & 98.47 & 99.86 & 98.08 & 97.52 & \textbf{95.53} \\
    \bottomrule
    \end{tabular}
    }
\end{table*}

\paragraph{Detailed Performance on ForenSynths.} \Cref{tab:forensyths} reports the cross-method generalization performance on ForenSynths~\cite{cvpr2020cnngenerated} across multiple GAN families, Deepfakes, and low-level/perceptual-loss manipulations using accuracy (ACC) and average precision (AP). RINE achieves the best average ACC (91.79\%), while WaveRep achieves the best average AP (98.87\%). Several prior methods exhibit high AP but significantly lower ACC on specific subsets, indicating inconsistent decision calibration under distribution shift. VINA achieves the second-best average ACC (86.99\%) and competitive AP (97.78\%), with strong performance on most GAN categories but a clear degradation on Deepfakes.

\begin{table*}[ht]
% \vspace{-0.3 cm}
    \centering
    \caption{\textbf{Benchmarking Results of Cross-method Evaluations in terms of ACC(\%) and AP(\%) Performance on the ForenSynth~\cite{cvpr2020cnngenerated} Dataset.} All evaluations used the authors' provided checkpoints and preprocessing pipelines.}
    \label{tab:forensyths}
\resizebox{\textwidth}{!}{
    \begin{tabular}{cc|cccccc|c|cc|cc|c}
    % \bottomrule 
    % \hline
\toprule
    \multirow{3}*{Methods} & \multirow{3}*{Type} & \multicolumn{6}{c|}{GAN} &  \multirow{3}*{\makecell[c]{Deep\\fakes}} & \multicolumn{2}{c|}{Low level} & \multicolumn{2}{c|}{Perceptual loss} & \multirow{3}{*}{AVG} \\
      % \multirow{3}*{Guided} & \multicolumn{3}{c|}{LDM} & \multicolumn{3}{c|}{Glide} &  \multirow{3}*{Dalle}  &  \multirow{3}*{mAP}\\ 
      
       % \cmidrule(lr){2-7} \cmidrule(lr){9-10}  \cmidrule(lr){11-12}  \cmidrule(lr){14-16}  \cmidrule(lr){17-19}  
    ~ & ~ & \makecell[c]{Pro-\\GAN} & \makecell[c]{Cycle-\\GAN} & \makecell[c]{Big-\\GAN} & \makecell[c]{Style-\\GAN}  & \makecell[c]{Gau-\\GAN}  & \makecell[c]{Star-\\GAN} & ~ & \makecell[c]{SITD}& \makecell[c]{SAN}& \makecell[c]{CRN}& \makecell[c]{IMLE} \\% ~ & {\makecell[c]{200\\steps}}& {\makecell[c]{200\\w/cfg}}& {\makecell[c]{100\\steps}}& {\makecell[c]{100\\27}} & {\makecell[c]{50\\27}} & \makecell[c]{100\\10} & ~ & ~\\
       % \midrule
       % \midrule
      % \hline
\midrule
    \multirow{2}*{\makecell{WaveRep\\~\cite{nips2025seeingwhatmatters}}} 
    & ACC & 92.99 & 72.67 & 73.28 & 90.80 & 70.16 & 94.50 & 87.64 & 84.44 & 80.59 & 58.54 & 58.54 & 78.56  \\
    ~ & AP & 99.85 & 98.53 & 96.41 & 97.56 & 99.61 & 99.87 & 96.65 & 99.68 & 99.41 & 100.0 & 100.0 & \textbf{98.87}  \\
\midrule
    \multirow{2}*{\makecell{RINE\\~\cite{eccv2024rine}}} 
    & ACC & 100.00 & 99.28 & 99.60 & 92.09 & 99.77 & 99.55 & 80.63 & 90.56 & 68.26 & 89.24 & 90.67 & \textbf{91.79} \\
    ~ & AP & 100.00 & 99.99 & 99.94 & 99.79 & 100.00 & 100.00 & 97.90 & 97.22 & 94.93 & 97.26 & 99.74 & \underline{98.80} \\
\midrule
    \multirow{2}*{\makecell{Effort (SD1.4)\\~\cite{icml2025effort}}} 
    & ACC & 83.62 & 92.51 & 86.83 & 80.42 & 82.14 & 98.95 & 69.40 & 58.33 & 52.74 & 51.91 & 51.74 & 73.51  \\
    ~ & AP & 92.59 & 94.46 & 86.12 & 86.72 & 80.65 & 99.89 & 75.22 & 62.16 & 52.91 & 68.32 & 62.27 & 78.30 \\
\midrule
    \multirow{2}*{\makecell{NPR\\~\cite{cvpr2024npr}}}
    & ACC & 99.79 & 91.29 & 82.12 & 96.75 & 78.31 & 85.99 & 79.44 & 58.89 & 65.30 & 50.00 & 50.00 & 76.17 \\
    & AP & 99.98 & 93.53 & 83.33 & 99.18 & 80.70 & 99.97 & 82.35 & 61.29 & 66.99 & 50.01 & 50.01 & 78.85 \\
\midrule
    \multirow{2}*{\makecell{DDA\\~\cite{nips2025dda}}}
    & ACC & 92.30 & 63.93 & 83.62 & 86.79 & 89.94 & 66.28 & 71.21 & 72.22 & 94.29 & 87.35 & 90.21 & 81.65  \\
    & AP & 99.04 & 81.23 & 94.31 & 94.71 & 96.91 & 69.06 & 82.61 & 70.77 & 98.80 & 93.94 & 98.25 & 89.06  \\
\midrule
    \multirow{2}*{\makecell{CO-SPY\\(ProGAN)}}
    & ACC &100.00 & 99.28 & 94.60 & 97.20 & 95.16 & 99.60 & 77.80 & 70.00 & 77.40 & 50.68 & 50.68 & 82.95 \\
    & AP &100.00 & 99.38 & 98.19 & 99.98 & 98.10 & 100.00 & 95.15 & 65.33 & 83.24 & 63.18 & 63.18 & 87.79  \\
\midrule
    \multirow{2}*{\makecell{CO-SPY\\~\cite{cvpr2025co-spy}}}
    & ACC & 74.25 & 57.87 & 71.25 & 61.50 & 69.26 & 62.06 & 65.42 & 51.94 & 71.92 & 54.07 & 44.84 & 62.22 \\
    & AP & 77.64 & 55.66 & 83.66 & 66.65 & 82.95 & 94.35 & 80.00 & 58.32 & 79.25 & 57.26 & 41.80 & 70.69 \\
\midrule
    \multirow{2}*{\makecell{VINA\\(140k+DDA)}} & ACC & 95.21 & 93.15 & 95.05 & 93.42 & 96.06 & 98.2 & 63.16 & 85.28 & 83.79 & 76.83 & 76.79 & \underline{86.99}   \\ 
    ~ & AP & 99.35 & 99.52 & 99.81 & 97.84 & 99.96 & 99.97 & 90.77 & 95.43 & 95.16 & 99.6 & 98.12 & 97.78  \\
%\hline
\bottomrule
    \end{tabular}
}
% \vspace{-0.5em}
\end{table*}

\paragraph{Detailed Performance on UniFD(LDM).} \Cref{tab:unifd} presents cross-method evaluations on the UniversalFakeDetect (LDM) dataset, comparing VINA against several state-of-the-art detection frameworks. VINA achieves the best average accuracy of 94.84\%, slightly surpassing the strongest baseline CO-SPY (ProGAN). It also obtains the best average AP of 99.01\%, demonstrating strong ranking robustness across LDM, GLIDE, and DALL-E subsets.

\begin{table*}[ht]
% \vspace{-0.3 cm}
    \centering
    \caption{\textbf{Benchmarking Results on UniversalFakeDetect (LDM)~\cite{cvpr2023unifd}: Accuracy and Average Precision.} All evaluations used the authors' provided checkpoints and preprocessing pipelines.}
    \label{tab:unifd}
\resizebox{\textwidth}{!}{
    \begin{tabular}{cc|cccccccc|c}
    % \bottomrule 
    % \hline
\toprule
    Method & Type & guided & ldm\_200 & \makecell{ldm\_\\200\_cfg} & ldm\_100 & \makecell{glide\_\\100\_27} & \makecell{glide\_\\50\_27} & \makecell{glide\_\\100\_10} & dalle & AVG \\
    % \multirow{3}*{Methods} & \multirow{3}*{Type} \multirow{3}*{Guided} & \multicolumn{3}{c|}{LDM} & \multicolumn{3}{c|}{Glide} &  \multirow{3}*{Dalle}  &  \multirow{3}*{mAP}\\ 

\midrule
    \multirow{2}*{\makecell{WaveRep\\~\cite{nips2025seeingwhatmatters}}} 
    & ACC & 79.40 & 78.80 & 81.45 & 76.70 & 79.45 & 80.85 & 82.85 & 84.10 & 80.45 \\
    ~ & AP & 88.93 & 92.66 & 95.57 & 90.66 & 93.17 & 94.32 & 96.81 & 98.53 & 93.83 \\
\midrule
    \multirow{2}*{\makecell{FatFormer\\~\cite{cvpr2024fatformer}}}
    & ACC & 76.05 & 96.70 & 94.85 & 98.60 & 94.30 & 94.60 & 94.15 & 98.70 & 93.49 \\
    & AP  & 91.92 & 99.53 & 99.22 & 99.89 & 99.27 & 99.50 & 99.33 & 99.84 & 98.56 \\
\midrule
    \multirow{2}*{\makecell{RINE\\~\cite{eccv2024rine}}}
    & ACC & 76.15 & 93.27 & 88.25 & 98.65 & 88.95 & 92.60 & 90.70 & 95.10 & 90.46 \\
    & AP  & 96.56 & 99.31 & 98.71 & 99.91 & 99.08 & 99.45 & 99.16 & 99.48 & \underline{98.96} \\
\midrule
    \multirow{2}*{\makecell{Effort (SD1.4)\\~\cite{icml2025effort}}} 
    & ACC & 78.05 & 93.80 & 91.70 & 96.45 & 93.05 & 91.60 & 92.05 & 96.65 & 91.67\\
    ~ & AP &84.34 & 98.57 & 97.83 & 99.35 & 98.53 & 98.02 & 98.30 & 99.41 & 96.79\\
\midrule
    \multirow{2}*{\makecell{NPR\\~\cite{cvpr2024npr}}}
    & ACC &74.40 & 96.67 & 96.60 & 96.15 & 96.20 & 96.80 & 96.95 & 88.70 & 92.81 \\
    & AP &79.51 & 98.91 & 98.85 & 98.87 & 98.91 & 98.92 & 98.99 & 96.78 & 96.22\\
\midrule
    \multirow{2}*{\makecell{DDA\\~\cite{nips2025dda}}}
    & ACC &90.45 & 84.12 & 84.20 & 83.95 & 75.25 & 74.70 & 79.85 & 72.05 & 80.57 \\
    & AP &98.17 & 99.26 & 99.36 & 99.11 & 88.22 & 87.42 & 92.88 & 84.28 & 93.59\\
\midrule
    \multirow{2}*{\makecell{CO-SPY\\(ProGAN)}}
    & ACC &79.45 & 95.88 & 94.55 & 97.55 & 97.00 & 97.85 & 97.85 & 94.85 & \underline{94.37}\\
    & AP &91.18 & 99.68 & 99.54 & 99.80 & 99.83 & 99.89 & 99.87 & 99.61 & 98.68\\
\midrule
    \multirow{2}*{\makecell{CO-SPY\\~\cite{cvpr2025co-spy}}}
    & ACC &62.40 & 83.53 & 84.85 & 82.00 & 72.85 & 68.20 & 75.40 & 81.30 & 79.01 \\
    & AP &85.93 & 89.29 & 91.00 & 86.83 & 77.91 & 74.33 & 81.40 & 87.13 & 84.23 \\
\midrule
    \multirow{2}*{\makecell{VINA\\(140k+DDA)}} & ACC & 75.80 & 99.30 & 99.30  & 99.30 & 94.45 & 96.10 & 96.55 & 97.95 & \textbf{94.84}  \\
    ~ & AP & 93.98  & 99.99  & 99.99  & 99.99  & 99.30  & 99.45  & 99.59  & 99.81  & \textbf{99.01} \\
%\hline
\bottomrule
    \end{tabular}
}
% \vspace{-0.5em}
\end{table*}

\paragraph{Detailed Performance on DiTFake.} \Cref{tab:ditfake} reports cross-method results on DiTFake~\cite{kdd2025safe} across FLUX, PixArt, and SD3. DDA achieves the best mACC (98.59), while VINA attains the best mAP (99.95) and reaches 100.0 ACC/AP on PixArt. Overall, VINA remains consistently high across generators with near-saturated AP.

\begin{table*}[ht]
% \vspace{-0.3 cm}
    \centering
    \caption{\textbf{Benchmarking Results of Cross-method Evaluations in terms of ACC(\%) and AP(\%) Performance on the DiTFake~\cite{kdd2025safe}}. $\dagger$ indicates that the results are obtained by us using the official pre-trained checkpoint and input data transformation.}
    \label{tab:ditfake}
\resizebox{\textwidth}{!}{
    \begin{tabular}{c|cccc|cccc}
    % \bottomrule 
    % \hline
\toprule
    Method & FLUX & PixArt & SD3 & mACC & FLUX & PixArt & SD3 & mAP \\
    % \multirow{3}*{Methods} & \multirow{3}*{Type} \multirow{3}*{Guided} & \multicolumn{3}{c|}{LDM} & \multicolumn{3}{c|}{Glide} &  \multirow{3}*{Dalle}  &  \multirow{3}*{mAP}\\ 

\midrule
    WaveRep~\cite{nips2025seeingwhatmatters} & 85.26 & 86.92 & 87.01 & 86.40 & 97.53 & 99.70 & 99.20 & 98.81 \\
    FatFormer~\cite{cvpr2024fatformer} & 54.49 & 67.16 & 58.34 & 60.00 & 61.85 & 79.11 & 69.12 & 70.03 \\
    RINE~\cite{eccv2024rine} & 56.27 & 59.59 & 55.12 & 56.99 & 92.29 & 93.43 & 91.62 & 92.45 \\
    Effort~\cite{icml2025effort} & 82.00 & 82.20 & 82.49 & 82.23 & 88.95 & 94.73 & 91.70 & 91.79 \\
    Co-SPY~\cite{cvpr2025co-spy} & 91.13 & 95.86 & 93.24 & 93.41 & 99.06 & 99.60 & 99.23 & 99.30 \\
    DDA~\cite{nips2025dda} & \textbf{97.15} & 99.47 & \textbf{99.14} & \textbf{98.59} & 99.71 & 99.99 & \textbf{99.96} & \underline{99.89} \\ 
\midrule
    VINA (140k+DDA) & 97.11 & 100.0 & 96.84 & \underline{97.98} & 99.95 & 100.0 & 99.90 & \textbf{99.95}  \\
%\hline
\bottomrule
    \end{tabular}
}
% \vspace{-0.5em}
\end{table*}

\paragraph{Detailed Performance on ARForensics.} \Cref{tab:arforensics} evaluates cross-method generalization on ARForensics~\cite{iccv2025d^3qe}, which consists of previously unseen auto-regressive generator architectures (e.g., Infinity, Janus Pro, VAR). VINA achieves the best overall performance, reaching 96.74 AVG ACC and 99.59 AVG AP. Compared to prior detectors, which exhibit larger accuracy drops on these out-of-distribution AR generators, the results indicate stronger robustness of unified training to novel generator architectures.

% \newcolumntype{Y}{>{\centering\arraybackslash}X}

\begin{table*}[ht]
% \vspace{-0.3 cm}
    \centering
    \caption{\textbf{ Benchmarking Results of Cross-method Evaluations in terms of ACC(\%) and AP(\%) Performance on the ARForensics~\cite{iccv2025d^3qe} Dataset.}  $\dagger$ indicates that the results are directly from their papers.}
    \label{tab:arforensics}
\resizebox{\textwidth}{!}{
    % \begin{tabular}{cc|ccccccc|c}
\begin{tabularx}{\textwidth}{cc *{8}{Y}}

\toprule
    Method & Metric & Infinity & Janus Pro & Llama Gen & MAR & \makecell{MAG\\ VIT2}& RAR & VAR  & AVG \\
\midrule
    \multirow{2}*{\makecell{D$^3$QE$\dagger$\\~\cite{iccv2025d^3qe}}}
    & ACC & 62.88 & 97.53 & 82.11 & - & 70.08 & 91.69 & 85.33 & 82.11 \\
    & AP  & 79.39 & 97.53 & 99.43 & - & 95.98 & 97.77 & 95.30 & 92.07 \\
\midrule
    \multirow{2}*{\makecell{FatFormer\\~\cite{cvpr2024fatformer}}} 
    & ACC & 53.67 & 84.88 & 69.73 & 87.33 & 73.00 & 98.12 & 87.01 & 79.11 \\
    ~ & AP & 57.84 & 93.36 & 83.68 & 95.20 & 86.07 & 99.81 & 95.66 & 87.37  \\
\midrule
    \multirow{2}*{\makecell{RINE\\~\cite{eccv2024rine}}} 
    & ACC & 72.91 & 84.30 & 71.74 & 83.42 & 67.77 & 99.03 & 73.72 & 78.98 \\
    ~ & AP & 95.57 & 98.22 & 96.53 & 98.09 & 93.26 & 99.96 & 96.34 & \underline{96.85} \\
\midrule
    \multirow{2}*{\makecell{Effort (SD1.4)\\~\cite{icml2025effort}}} 
    & ACC & 86.84 & 87.69 & 79.36 & 77.42 & 73.30 & 85.92 & 81.55 & 81.73 \\
    ~ & AP & 94.06 & 98.46 & 88.18 & 83.96 & 80.23 & 90.41 & 90.58 & 89.41  \\
\midrule
    \multirow{2}*{\makecell{CO-SPY (SD1.4)\\~\cite{cvpr2025co-spy}}}
    & ACC & 93.09 & 79.68 & 60.10 & 65.88 & 59.21 & 64.98 & 63.12 & 69.44  \\
    & AP & 99.16 & 95.89 & 83.93 & 89.18 & 83.09 & 88.93 & 88.61 & 89.83 \\
\midrule
    \multirow{2}*{\makecell{CO-SPY (ProGAN)\\~\cite{cvpr2025co-spy}}}
    & ACC & 93.67 & 79.47 & 76.08 & 84.23 & 77.68 & 94.16 & 94.11 & 85.63 \\
    & AP & 98.21 & 87.90 & 87.38 & 90.84 & 87.42 & 98.60 & 98.65 & 92.71 \\
\midrule
    \multirow{2}*{\makecell{DDA\\~\cite{nips2025dda}}}
    & ACC & 98.04 & 98.47 & 93.38 & 93.83 & 67.31 & 85.83 & 75.78 & \underline{87.52} \\
    & AP & 99.84 & 99.90 & 98.66 & 99.03 & 88.40 & 97.51 & 92.96 & 96.61 \\
\midrule
    \multirow{2}*{\makecell{VINA\\(140k+DDA)}} & ACC & 99.84 & 99.90 & 98.07 & 99.01 & 84.08 & 98.17 & 98.11 & \textbf{96.74} \\
    ~ & AP & 100.0 & 100.0 & 99.77 & 99.95 & 97.56 & 99.96 & 99.92 & \textbf{99.59} \\
%\hline
\bottomrule
    \end{tabularx}
}
% \vspace{-0.5em}
\end{table*}

\paragraph{Detailed Performance on Magic Videos.} The Magic Videos dataset primarily comprises high-resolution videos generated by state-of-the-art commercial models, presenting a significant challenge for current detection frameworks.
As shown in \cref{tab:magic-acc}, existing AI image and video detection methods struggle to maintain robust performance on this benchmark, with most baselines failing to achieve satisfactory results.
In contrast, VINA demonstrates superior generalization, achieving the best average accuracy (77.08\%) and average precision (95.10\%).

\begin{table}[h]
    \centering
    \caption{\textbf{Benchmarking Results in terms of ACC and AP Performance on Magic Videos Testset~\cite{iclr2026preserving}.}}
    \label{tab:magic-acc}
    \resizebox{\textwidth}{!}{
    \begin{tabular}{cc|ccccccc}
    \toprule
        {Model} & Metric & Wan 2.1 & Wan-1.3B & Hailuo & Seaweed & Seedance & StepVideo & AVG \\ 
        % ~ & ~ & ACC & AP & ACC & AP & ACC & AP & ACC & AP & ACC & AP \\ 
\midrule
    \multirow{2}*{\makecell{WaveRep\\~\cite{nips2025seeingwhatmatters}}} 
    & ACC & 56.98 & 53.77 & 58.84 & 58.84 & 56.74 & 58.60 & 57.30 \\
    ~ & AP & 82.13 & 86.22 & 94.93 & 99.16 & 80.38 & 91.10 & \underline{88.99} \\
\midrule
    \multirow{2}*{\makecell{FatFormer\\~\cite{cvpr2024fatformer}}} 
    & ACC & 50.00 & 46.75 & 50.00 & 50.00 & 50.00 & 50.00 & 49.46  \\
    ~ & AP & 38.95 & 34.66 & 41.22 & 39.97 & 36.89 & 48.99 & 40.11 \\
\midrule
    \multirow{2}*{\makecell{RINE\\~\cite{eccv2024rine}}}
    & ACC & 40.47 & 41.27 & 40.70 & 48.14 & 39.53 & 45.81 & 42.65 \\
    & AP & 32.62 & 31.12 & 32.79 & 47.24 & 31.35 & 42.01 & 36.19 \\
\midrule
    \multirow{2}*{\makecell{NPR\\~\cite{cvpr2024npr}}}
    & ACC & 50.23 & 47.60 & 50.23 & 50.93 & 50.23 & 50.00 & 49.87 \\
    & AP & 64.86 & 40.01 & 56.91 & 87.61 & 75.41 & 87.88 & 68.78 \\
\midrule
    \multirow{2}*{\makecell{Effort\\~\cite{icml2025effort}}}
    & ACC & 83.95 & 35.79 & 69.30 & 81.63 & 77.21 & 56.51 & 67.40 \\
    & AP & 90.87 & 35.89 & 78.97 & 87.96 & 87.47 & 63.14 & 74.05 \\
\midrule
    \multirow{2}*{\makecell{Co-SPY\\~\cite{cvpr2025co-spy}}}
    & ACC & 68.84 & 59.93 & 63.26 & 64.65 & 57.91 & 77.67 & 65.38 \\
    & AP & 75.40 & 62.44 & 70.39 & 71.99 & 59.55 & 90.42 & 71.70 \\
\midrule
    \multirow{2}*{\makecell{DDA\\~\cite{nips2025dda}}}
    & ACC & 66.28 & 79.28 & 77.44 & 85.81 & 62.33 & 69.30 & \underline{73.41} \\
    & AP & 76.78 & 94.22 & 90.14 & 97.90 & 71.62 & 79.73 & 85.07 \\
\midrule
    \multirow{2}*{\makecell{VINA\\(140k+DDA)}}
    & ACC & 75.35 & 89.90 & 74.19 & 73.02 & 74.42 & 75.58 & \textbf{77.08}  \\
    & AP & 95.44 & 98.07 & 94.47 & 90.51 & 95.00 & 97.09 & \textbf{95.10}  \\
\bottomrule
    \end{tabular}

    }
    % \vspace{-3mm}
\end{table}

% \FloatBarrier
% \newpage

\paragraph{Detailed Performance on GenVideo-Val.} As observed in \cref{tab:genvideo_full}, most image-based detectors (e.g., NPR, FatFormer) fail significantly on the GenVideo-Val benchmark~\cite{arxiv2024demamba}, whereas DDA maintains high recall but suffers from low F1 and AP scores. This discrepancy suggests that image detectors are still capable of perceiving forgery artifacts, yet their decision thresholds are severely disrupted by the inherent domain gap and the lower noise levels characteristic of video data. By introducing video samples as physically augmented regularization, VINA effectively recalibrates the discriminative boundary, achieving the best comprehensive performance with 93.20\% average recall, 81.52\% average F1, and 92.83\% average AP.

\begin{table}[h]
    \caption{\textbf{Benchmarking Evaluation in terms of Recall, F1 score (F1), and average precision (AP) on GenVideo-Val~\cite{arxiv2024demamba}.} $\dagger$ indicates that the results are directly from their papers.}
    \label{tab:genvideo_full}
    \centering
    \resizebox{\textwidth}{!}{
    \begin{tabular}{ccccccccccccc}
    \toprule
   \multirow{2}{*}{Model} & \multirow{2}{*}{Metric}&\multirow{2}{*}{Sora}&Morph&\multirow{2}{*}{Gen2}&\multirow{2}{*}{HotShot}&\multirow{2}{*}{Lavie}&\multirow{2}{*}{Show-1}&Moon&\multirow{2}{*}{Crafter}&Model&Wild&\multirow{2}{*}{Avg.}\\
   &&&Studio&&&&&Valley&&Scope&Scrape\\
   \midrule
   \multirow{3}{*}{\makecell{UNITE\\~\cite{cvpr2025unite}}}
        & Recall & 92.11 & 100.0 & 94.62 & 96.93 & 98.12 & 99.86 & 98.69 & 100.0 & 96.29 & 89.89 & 89.60 \\
        & F1 & - & - & - & - & - & - & - & - & - & - & - \\
        & AP & 88.57 & 100.0 & 100.0 & 90.16 & 89.91 & 98.34 & 99.52 & 100.0 & 98.96 & 92.56 & \underline{92.76} \\
    \midrule
    \multirow{3}{*}{\makecell{WaveRep\\~\cite{nips2025seeingwhatmatters}}}
        & Recall & 87.50 & 99.57 & 100.00 & 80.00 & 95.21 & 97.71 & 100.00 & 99.93 & 97.57 & 81.93 & 87.50\\
        & F1 & 7.78 & 54.62 & 70.50 & 46.38 & 68.57 & 53.88 & 52.01 & 70.73 & 53.82 & 53.39 & 53.17 \\
        & AP & 41.93 & 99.20 & 99.93 & 67.55 & 92.95 & 93.85 & 99.70 & 99.71 & 94.78 & 77.39 & 86.70 \\
    \midrule
    \multirow{3}{*}{\makecell{NPR\\~\cite{cvpr2024npr}}}
        & Recall & 3.57  & 7.00  & 2.83  & 1.29  & 2.00  & 0.00  & 0.32  & 3.00  & 16.71  & 3.35 & 4.01 \\
        & F1 & 5.33  & 12.79  & 5.43  & 2.48  & 3.88  & 0.00  & 0.62  & 5.77  & 28.06  & 6.37 & 7.07 \\
        & AP & 2.06  & 39.20  & 27.84  & 7.06  & 16.73  & 7.31  & 8.74  & 45.43  & 45.78  & 23.09 & 22.32 \\
    \midrule
    \multirow{3}{*}{\makecell{FatFormer\\~\cite{cvpr2024fatformer}}}
        & Recall & 1.79  & 2.57  & 1.23  & 0.14  & 2.36  & 4.14  & 0.00  & 4.01  & 37.00  & 11.02 & 6.43 \\
        & F1 & 3.08  & 4.96  & 2.42  & 0.28  & 4.58  & 7.87  & 0.00  & 7.66  & 53.57  & 19.69 & 10.41 \\
        & AP & 0.79  & 22.45  & 15.46  & 8.33  & 24.48  & 18.87  & 5.80  & 31.26  & 64.30  & 40.51 & 23.23 \\
    \midrule
    \multirow{3}{*}{\makecell{RINE\\~\cite{eccv2024rine}}}
        & Recall & 28.57  & 5.86  & 1.01  & 2.29  & 5.14  & 3.14  & 2.40  & 5.51  & 23.43  & 7.78 & 8.51 \\
        & F1 & 42.11  & 11.01  & 2.00  & 4.44  & 9.76  & 6.06  & 4.65  & 10.41  & 37.79  & 14.37 & 14.26 \\
        & AP & 64.76  & 65.22  & 63.58  & 36.26  & 62.23  & 48.29  & 46.74  & 77.39  & 86.18  & 63.24 & 61.39 \\
    \midrule
    \multirow{3}{*}{\makecell{Effort\\~\cite{icml2025effort}}}
        & Recall & 33.93 & 68.29 & 56.16  & 29.86  & 51.79  & 40.14  & 75.24  & 63.88  & 77.86 & 49.78 & 54.69 \\
        & F1 & 25.17 & 76.24  & 69.48  & 42.44  & 65.88  & 53.17  & 80.31  & 75.45  & 82.51  & 63.01 & \underline{63.37} \\
        & AP & 30.81  & 87.26  & 89.00  & 59.99  & 84.29  & 71.67  & 89.36  & 90.56  & 91.56  & 77.43 & 77.19 \\
    \midrule
    \multirow{3}{*}{\makecell{Co-SPY\\~\cite{cvpr2025co-spy}}}
        & Recall & 48.21  & 96.00  & 93.62  & 18.71  & 53.71  & 71.29  & 97.92  & 90.77  & 91.71  & 72.94 & 73.49 \\
        & F1 & 3.00  & 43.52  & 58.89  & 10.29  & 38.88  & 34.24  & 41.49  & 57.91  & 41.99  & 40.68 & 37.09 \\
        & AP & 1.97  & 76.57  & 79.94  & 7.96  & 37.70  & 37.98  & 78.24  & 76.67  & 68.19  & 51.72 & 51.69\\
    \midrule
    \multirow{3}{*}{\makecell{DDA\\~\cite{nips2025dda}}}
        & Recall & 51.79  & 97.29  & 98.77  & 98.86  & 97.64  & 98.86  & 97.92  & 97.85  & 90.43  & 82.03 & \underline{91.14} \\
        & F1 & 0.89  & 17.37  & 29.62  & 17.63  & 29.63  & 17.63  & 15.92  & 29.66  & 16.25  & 18.62 & 19.32 \\
        & AP & 0.54  & 65.57  & 91.55  & 80.03  & 84.67  & 85.75  & 81.24  & 84.50  & 58.62  & 52.69 & 68.52 \\
    \midrule
    \multirow{3}{*}{\makecell{VINA\\(140k+DDA)}}
        & Recall & 75.00 & 99.86 & 100.0 & 94.29 & 97.71 & 96.29 & 100.0 & 99.79 & 98.71 & 70.35 & \textbf{93.20} \\
        & F1 & 29.79 & 88.31 & 93.75 & 85.49 & 92.68 & 86.52 & 87.19 & 93.72 & 87.75 & 73.95 & \textbf{81.52} \\ 
        & AP & 56.82 & 99.74 & 99.96 & 95.90 & 99.03 & 97.53 & 99.97 & 99.87 & 99.08 & 80.42 & \textbf{92.83} \\
    \bottomrule
    \end{tabular}}
    % \vspace{-5mm}
\end{table}

\section{More Ablation Studies}
\label{sec:appendix_ablation}

\begin{figure}[ht]
    \centering
    \includegraphics[width=0.52\linewidth]{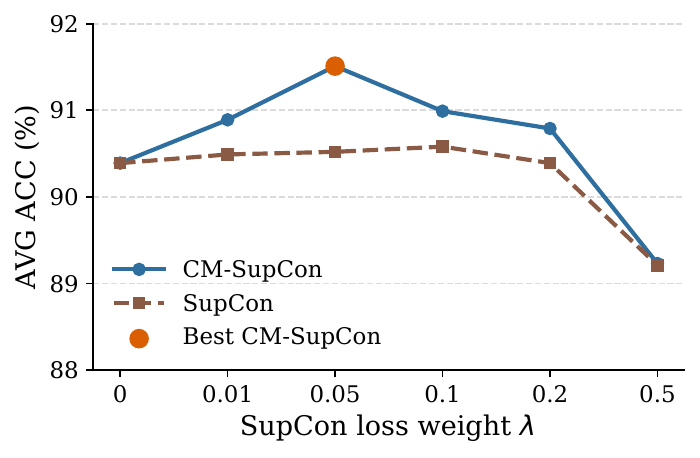}
    \caption{\textbf{Ablation Study on SupCon Loss Weight.} We vary the supervised contrastive loss weight $\lambda$ and report AVG ACC, computed as the average accuracy over all 14 benchmarks. CM-SupCon defines positives across modalities with the same real/fake label, whereas vanilla SupCon uses all same-label samples as positives. CM-SupCon achieves the best result at $\lambda=0.05$.}
    \label{fig:supcon_lambda_ablation}
\end{figure}

This section includes additional ablation results in \cref{tab:ablation_source,tab:ablation_tuning,tab:ablation_processing,fig:supcon_lambda_ablation}.

\begin{table}[t]
    \caption{\textbf{Ablation study on training data source.} Video Data All = 140k, Image Data All = GenImage SD1.4 Subset. Experiments use a DINOv3-L backbone without SupCon loss.}
    \label{tab:ablation_source}
    \centering
    % \resizebox{\linewidth}{!}{
    \begin{tabular}{cc|cccc}
    \toprule
    \makecell{Video Data\\(140k)} & \makecell{Image Data\\(GenImage)} & \makecell{Video\\AVG} & \makecell{Image\\AVG} & \makecell{Chame\\leon} & \makecell{Mean\\ACC} \\
    \midrule
    ALL & - & \underline{88.74} & 79.77 & 49.45 & 72.65   \\
    ALL & Real & 85.51 & 78.35 & \underline{89.47} & \underline{84.44}  \\
    ALL & Fake & 86.08 & 74.28 & 47.69 & 69.35  \\
    \midrule
    - & ALL & 59.60 & 75.08 & 59.49 & 64.72  \\
    Real & ALL & 55.64 & 79.69 & 60.58 & 65.30 \\
    Fake & ALL & 64.63 & \underline{83.41} & 88.03 & 78.69 \\
    \midrule
    ALL & ALL & \textbf{89.79} & \textbf{87.31} & \textbf{91.48} & \textbf{89.53}  \\
    \bottomrule
    \end{tabular}
    % }
\end{table}

\paragraph{Ablation Study on Data Sources.} \Cref{tab:ablation_source} uses a DINOv3-L backbone without SupCon loss to isolate the effect of training data composition. Performance improves substantially when video data is supplemented with real images, or when image data is supplemented with synthetic videos. The best mean accuracy is achieved when the complete dataset containing both modalities is used, highlighting the positive contribution of diverse video and image sources.

\paragraph{Ablation Study on Tuning Method.} \Cref{tab:ablation_tuning} compares tuning strategies on the same 140k+DDA training set. Linear probing underperforms substantially (Mean 76.9), especially on Chameleon, while LoRA yields large gains across all evaluation groups and improves the mean accuracy to 89.9. End-to-end fine-tuning achieves the best overall mean accuracy (92.0), with the strongest video and image performance.

\begin{table}[ht]
    \centering
    \caption{\textbf{Ablation Study on Tuning Methods.} All methods are DINOv3-L backbones trained on 140k+DDA dataset. For LoRA, lr=1e-5 and for LP, lr=1e-4.}
    \label{tab:ablation_tuning}
    \begin{tabular}{l|cccc}
    \toprule
    Method & Video ACC & Image ACC & Chameleon & Mean \\
    \midrule
    LP & 82.2 & 83.2 & 65.4 & 76.9 \\
    LoRA (r=16) & 89.7 & 89.0 & 91.0 & \underline{89.9} \\
    Full (Ours) & \textbf{90.4} & \textbf{94.1} & \textbf{91.4} & \textbf{92.0} \\
    \bottomrule
    \end{tabular}
    \vspace{-5mm}
\end{table}

\begin{table}[ht]
    \centering
    \caption{\textbf{Ablation Study on Training Image Transform Methods.} All methods use a DINOv3-L backbone trained on the 140k+DDA dataset without SupCon loss, with target resolution 256$\times$256.}
    \begin{tabular}{l|cccc}
    \toprule
    Method & Video ACC & Image ACC & Chameleon & Mean \\
    \midrule
    resize & 88.13 & 92.59 & 90.36 & 90.36 \\
    center crop & 87.27 & 93.93 & 79.97 & 87.06 \\
    random crop & 86.38 & 96.31 & 82.73 & 88.47 \\
    random resized crop & 88.45 & 95.57 & 88.10 & \underline{90.71}\\
    shorter resize + random crop (Ours) & 90.29 & 92.96 & 90.98 & \textbf{91.41} \\
    \bottomrule
    \end{tabular}
    
    \label{tab:ablation_processing}
\end{table}

\paragraph{Ablation Study on Image Transforms.} \Cref{tab:ablation_processing} studies the effect of input transformations using a DINOv3-L detector trained on the 140k+DDA dataset without SupCon loss. Standard cropping introduces a clear trade-off across test settings: for example, center crop improves Image accuracy but substantially degrades Chameleon accuracy, indicating reduced robustness to variations in resolution and framing. The shorter-resize plus random-crop strategy achieves the best mean score (91.41) and the highest Video accuracy (90.29), demonstrating more consistent performance across diverse high- and variable-resolution inputs.

\paragraph{Ablation Study on SupCon Loss Weight.} \Cref{fig:supcon_lambda_ablation} evaluates the sensitivity to the supervised contrastive loss weight $\lambda$ using the same DINOv3-L model trained on the 140k+DDA dataset. Here, AVG ACC denotes the average accuracy over all 14 benchmarks used in our evaluation. We compare our CM-SupCon loss with a vanilla SupCon variant that treats all samples with the same real/fake label as positives, regardless of modality. CM-SupCon obtains a clear gain over the no-SupCon baseline and reaches the best result at $\lambda=0.05$ (91.51), while vanilla SupCon provides only marginal improvement and peaks at $\lambda=0.1$ (90.58). Larger weights reduce performance for both variants, especially when $\lambda=0.5$, indicating that excessive contrastive regularization can over-constrain the representation and interfere with binary classification.

\section{Scaling Potentials}
\label{sec:appendix_scaling}

This section includes temporal scaling results in \cref{tab:scaling_temporal}.

\begin{table}[t!]
    \centering
    \caption{\textbf{Test-time temporal scaling.} We vary the number of uniformly sampled frames per video at inference (T) while keeping the DINOv3-L model trained without SupCon loss fixed; logits from multiple frames are averaged.}
    \begin{tabular}{l|ccccc}
    \toprule
    Sampled Frame & Magic & GenVideo & GenBuster++ & DeepTraceReward & Mean \\
    \midrule
    T=1 & 80.18 & 96.78 & 86.35 & 97.85 & 90.29 \\
    T=2 & 79.89 & 97.01 & 87.3 & 98.16 & \textbf{90.59} \\
    T=4 & 79.14 & 97.23 & 87.45 & 98.33 & \underline{90.54} \\
    T=8 & 77.67 & 97.35 & 88.05 & 98.46 & 90.38 \\
    \bottomrule
    \end{tabular}
    \label{tab:scaling_temporal}
    \vspace{-3mm}
\end{table}

We investigate whether the same trained model can leverage longer video context by simply increasing the number of sampled frames at inference, without modifying training. Unless otherwise specified, these experiments use DINOv3-L models trained without SupCon loss.

\paragraph{Scaling Temporal Frames.} \Cref{tab:scaling_temporal} probes temporal scalability by increasing the number of sampled frames per video only at test time (from $T=1$ to $T=8$), without changing training. Performance remains stable and slightly improves when using more frames, with mean ACC peaking at 90.59 for $T=2$. In future work, we will explore increasing the number of sampled frames during training and explicitly modeling temporal dynamics to further benefit from longer temporal context.

% Ablation Resolution Processing
% Ablation on tuning method
% temporal scaling
\FloatBarrier

\section{Discussion}
\label{sec:appendix_discussion}

\paragraph{Broader Impacts.} This work focuses on detecting AI-generated images and videos, rather than introducing or improving AIGC generation techniques. By improving cross-modal detection robustness, VINA can help reduce misunderstanding and misinformation caused by synthetic media, supporting content verification, platform moderation, and forensic analysis.

At the same time, detection research can have dual-use risks. Insights into detector behavior may be used to improve generation pipelines or post-processing strategies that evade current detectors. In practical deployment, another risk is over-confidence: no detector is perfect, and false positives or false negatives may lead to unfair moderation decisions, misplaced trust, or incorrect attribution. We therefore recommend using AIGC detectors as decision-support tools together with provenance metadata, human review, and transparent uncertainty reporting, rather than as the sole basis for high-stakes judgments.

\end{document}